\ifcsname iflatexml\endcsname
\documentclass{article}
\usepackage[numbers,sort&compress]{natbib}
\usepackage{authblk,xcolor,colortbl,booktabs,graphicx,caption,wrapfig,changepage,enumitem,url,hyperref,amsfonts}
\else
\documentclass[9pt,twocolumn,twoside]{osajnl}
\fi

\usepackage{times}
\usepackage{amsmath}
\usepackage{amssymb}
\usepackage{pifont}
\usepackage{times}
\usepackage{epsfig}
\usepackage{graphicx}
\usepackage{float}
\usepackage{wrapfig}

\usepackage{amsmath,amssymb}

\usepackage{bm,xspace}
\usepackage{comment}
\usepackage{verbatim}
\usepackage{multirow}
\usepackage{multicol}
\usepackage{balance}
\usepackage{url}
\usepackage{booktabs}
\usepackage{etoolbox,siunitx}
\ifcsname iflatexml\endcsname\else\usepackage{calc}\fi
\usepackage{pifont,hologo}
\usepackage{color}

\usepackage[super]{nth}
\usepackage{nicefrac}
\def\aa{\mathbf{a}}

\def\ss{\mathbf{s}}
\def\tt{\mathbf{t}}

\def\AA{\mathbf{A}}

\DeclareMathSymbol{@}{\mathord}{letters}{"3B}

\newcommand\mypara[1]{\paragraph{#1.}}

\def\latex/{\LaTeX}
\def\bibtex/{\hologo{BibTeX}}

\usepackage[official]{eurosym}

\usepackage{subcaption}

\usepackage{hyperref}
\hypersetup{
    colorlinks=true,
    linkcolor=magenta,
    filecolor=magenta,      
    urlcolor=blue,
    pdftitle={On Agentic Manipulation Policies},
    pdfpagemode=FullScreen,
    }

\usepackage[capitalize]{cleveref}
\Crefname{table}{Table}{Tables}
\crefname{table}{Tab.}{Tabs.}
\crefname{section}{Sec.}{Secs.}
\Crefname{section}{Section}{Sections}
\Crefname{figure}{Figure}{Figures}
\crefname{figure}{Fig.}{Figs.}
\crefname{algorithm}{Alg.}{Algs.}
\Crefname{algorithm}{Algorithm}{Algorithms}
\crefname{equation}{Eq.}{Eqs.}
\Crefname{equation}{Equation}{Equations}

\ifcsname iflatexml\endcsname\else
\journal{ol}
\dates{\makebox[\textwidth][c]{\rule{0pt}{13pt}\rule[-20.55pt]{0pt}{20.55pt}\normalfont\sffamily\bfseries\Large\href{https://apple-aiml-research.github.io/ml-amp/}{https://apple-aiml-research.github.io/ml-amp/}}}
\fi
\renewcommand{\thesubsection}{\thesection\Alph{subsection}}
\ifcsname iflatexml\endcsname\else
\setboolean{shortarticle}{false}
\fi
\newcommand{\authbreak}{,\ }
\newcommand{\affilsize}{\normalsize}
\newcommand{\references}{\bibliography{bibliography}}
\newenvironment{widefigure}{\begin{figure*}}{\end{figure*}}

\usepackage{graphicx}

\date{}

\usepackage[nohyperlinks, printonlyused, nolist]{acronym}

\usepackage{url}
\usepackage{siunitx}
\usepackage{verbatim}
\usepackage{booktabs}
\usepackage{multirow}

\newcommand{\sysname}{AMP1\xspace}
\newcommand{\sysnameexp}{AMP1+\xspace}

\definecolor{cP}{HTML}{D70015}    %
\definecolor{cNP}{HTML}{0040DD}   %
\definecolor{cTU}{HTML}{248A3D}   %
\definecolor{cSS}{HTML}{D70015}   %
\definecolor{cMS}{HTML}{0040DD}   %
\definecolor{cSR}{HTML}{D70015}   %
\definecolor{cSG}{HTML}{0040DD}   %
\definecolor{cEX}{HTML}{248A3D}   %
\definecolor{cAF}{HTML}{8944AB}   %
\definecolor{cOR}{HTML}{FF9500}   %
\definecolor{cME}{HTML}{6C6C70}   %
\definecolor{cPNP}{HTML}{8944AB}  %

\newcommand{\badge}[2]{{\setlength{\fboxsep}{1.2pt}\colorbox{#1}{\textcolor{white}{\scriptsize\textbf{#2}}}}}
\newcommand{\tP}{\badge{cP}{P}}          %
\newcommand{\tNP}{\badge{cNP}{NP}}       %
\newcommand{\tSS}{\badge{cSS}{SS}}       %
\newcommand{\tMS}{\badge{cMS}{MS}}       %
\newcommand{\tSR}{\badge{cSR}{SR}}       %
\newcommand{\tSG}{\badge{cSG}{SG}}       %
\newcommand{\tEX}{\badge{cEX}{EX}}       %
\newcommand{\tOR}{\badge{cOR}{OR}}       %
\newcommand{\tAF}{\badge{cAF}{AF}}       %
\newcommand{\tME}{\badge{cME}{ME}}   %

\def\scititle{
On Agentic Manipulation Policies
}

\title{\bfseries \boldmath \scititle}

\ifcsname iflatexml\endcsname
\author{
    Omid~Bagherieh$^\dagger$,
    Naveen~Nikamanth~Appiah~Balaji$^\dagger$,
    Chen~Chen$^\dagger$,
    Jesse~T.~Gonzalez$^\dagger$,
    Negin~Heravi$^\dagger$,
    Fahad~Islam$^\dagger$,
    Xueting~Li$^\dagger$,
    Jean-Baptiste~Regli$^\dagger$,
    Kejia~Ren$^\dagger$,
    Petr~\v{S}vec$^\dagger$,
    David~J.~Yoon$^\dagger$,
    Matthias~M\"uller,
    Vladlen~Koltun\\
    Apple\\
    $^\dagger$These authors contributed equally to this work. Alphabetical order.\\
    \phantom{.}
}
\else
\author{
	Omid~Bagherieh$^\dagger$,
    Naveen~Nikamanth~Appiah~Balaji$^\dagger$,
    Chen~Chen$^\dagger$,
    Jesse~T.~Gonzalez$^\dagger$\authbreak
    Negin~Heravi$^\dagger$,
    Fahad~Islam$^\dagger$,
    Xueting~Li$^\dagger$,
    Jean-Baptiste~Regli$^\dagger$\authbreak
    Kejia~Ren$^\dagger$,
    Petr~\v{S}vec$^\dagger$,
    David~J.~Yoon$^\dagger$\authbreak
    Matthias~M\"uller,
    Vladlen~Koltun\\
  	\affilsize Apple\\
  	\affilsize $^\dagger$These authors contributed equally to this work. Alphabetical order.
}
\fi
\ifcsname iflatexml\endcsname\makeatletter\@add@frontmatter{ltx:subtitle}{\bfseries\href{https://apple-aiml-research.github.io/ml-amp/}{https://apple-aiml-research.github.io/ml-amp/}}\makeatother\fi

\makeatletter
\newcommand{\lxtop}[1]{\ifcsname iflatexml\endcsname\fpeval{#1-13/\strip@pt\unitlength}\else#1\fi}
\makeatother

\begin{document}

\begin{abstract}
Recent work on robot manipulation has focused on specialist policies trained on demonstrations. We show that generalist AI models can control manipulators more reliably than specialist policies such as vision-language-action models (VLAs) and world-action models (WAMs). We systematically analyze agentic manipulation policies (AMPs), in which a generalist AI model commands the manipulator directly, adapting its actions as the task unfolds. We present AMP1, a reference harness that runs unchanged across single-arm, bimanual, and mobile manipulation. AMP1 achieves state-of-the-art results on the RoboLab-120 and CaP-Gym benchmarks, with no demonstrations or task-specific training. On RoboLab-120, AMP1 reaches an 85.6\% success rate zero-shot, compared to 49.2\% and 42.9\% for leading VLA and WAM policies. The agent succeeds zero-shot in 104 of 108 real-world trials spanning twelve tasks and three robot instances, reducing failure rates by 20x and 15x relative to strong VLA and WAM baselines. The basic harness supplies no grasping, pushing, or placing skills; the agent develops its own strategies, such as searching for hidden objects, using objects as tools, and calibrating its cameras.  With the addition of a persistent memory store, the agent can also learn from experience: an augmented harness with a memory representation that is carried across episodes lowers the failure rate on RoboDojo bimanual manipulation tasks by 4.6x relative to zero-shot deployment. Our analysis suggests that AMPs are substantially more general and robust than specialist manipulation policies, providing a promising foundation for robot manipulation.
\end{abstract}

\maketitle

\begin{figure*} %
	\centering
	\includegraphics[width=\linewidth]{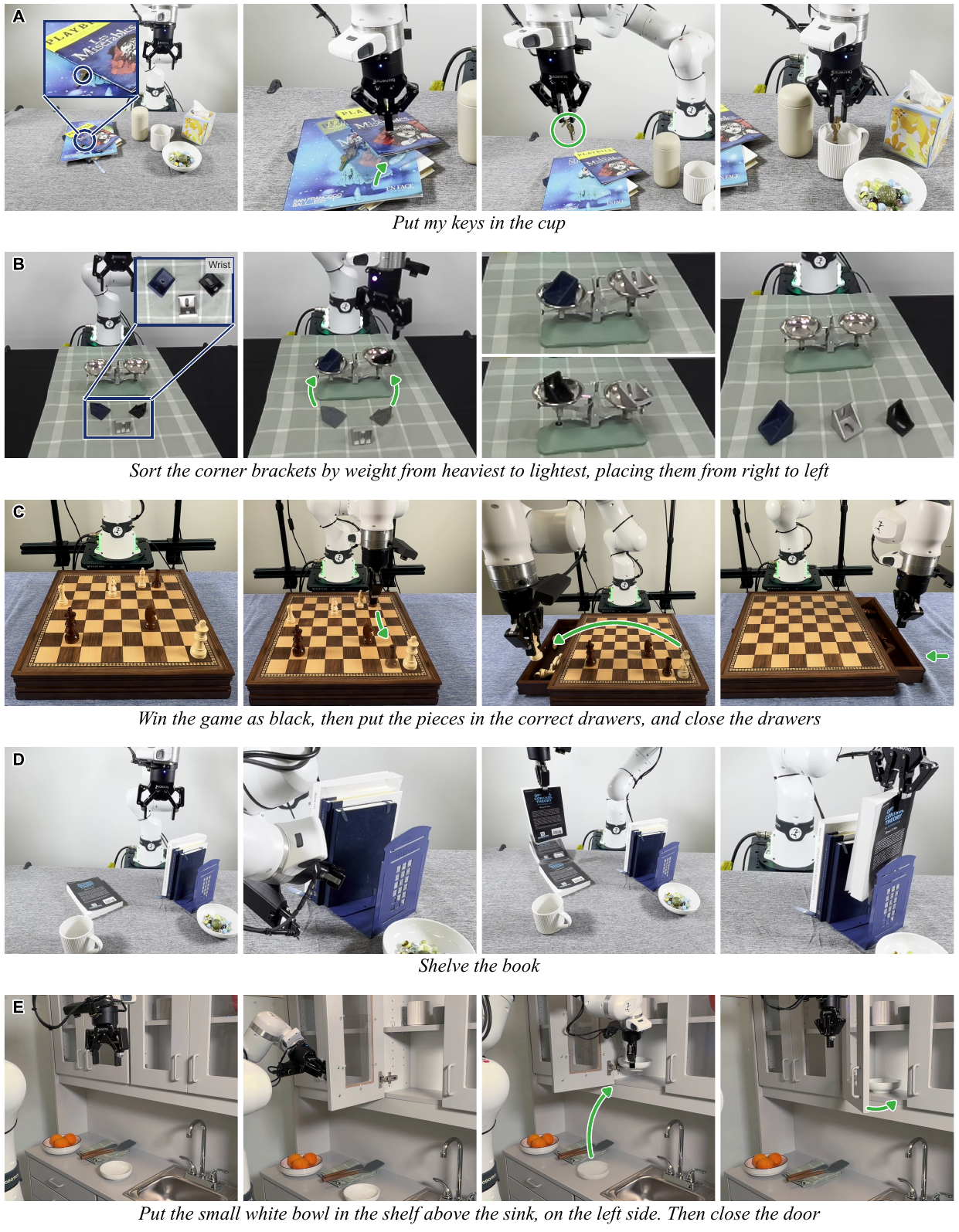}
	\caption{\textbf{Zero-shot real-world manipulation with \sysname.} \sysname solves diverse, long-horizon manipulation tasks directly in the real world from a natural language instruction without task-specific fine-tuning or predefined skills. It devises complex strategies, such as (\textbf{A}) searching for hidden objects, (\textbf{B}) using a balance scale as a measuring instrument,
    (\textbf{C}) planning and executing a winning strategy in a game of chess, (\textbf{D}) combining non-prehensile and prehensile actions where a direct grasp is infeasible, and (\textbf{E}) manipulating articulated objects such as a kitchen cabinet door. Task instructions given to \sysname for each episode are shown below the corresponding image sequence.}
	\label{fig:first}
\end{figure*}

\section{Introduction}
\label{sec:introduction}

In nature, manipulation draws on general intelligence. Animals can recruit their memory, reasoning, and problem-solving faculties when interacting with objects, with cognitive processes continuing to shape action as it unfolds~\cite{BirdEmery2009,CisekKalaska2010,Duncan2010}. In robotics, much recent work has focused on models specialized for robot control, including vision-language-action models (VLAs) and world-action models (WAMs), trained or fine-tuned on robot demonstrations~\cite{Brohan2023,Ye2026}.

Generalist AI models have also been applied to manipulation, often through skill orchestration or offline program synthesis. In the first setting (skill orchestration), the model selects and sequences low-level skills such as grasping and placing~\cite{Ahn2022,Huang2022,Li2026:RoboClaw}. It can revise the plan as new observations arrive, but the available skills limit the generalist model's control authority and the granularity at which it can command the manipulator. Since the generalist model's interactions with the world are mediated by the provided skills, it is constrained by the skills' limitations. In the second setting (offline program synthesis, sometimes known as `code-as-policies'), the model writes a control program before execution~\cite{Liang2023,rho,Chen2026:GaP,Shen2026:EmbodiedSWE,harnesspai}. The program can incorporate perception and feedback, but the generalist model is not in the loop and cannot reconsider the overall control strategy as the task unfolds.

A key obstacle to direct manipulation has been the limited spatial cognition of generalist frontier models~\cite{Ramakrishnan2025}. However, spatial cognition has been advancing~\cite{Chen2026:BabyVision,Zeller2026}, with recent models demonstrating convincing spatial planning, sensorimotor coordination, and embodied fluid intelligence~\cite{Anonymous2026:SpatialCognition,anonymous2026benchmarking}. These advances make a third approach increasingly viable: the generalist model remains in the control loop and commands the manipulator directly, revising its actions as the task unfolds~\cite{capx,Hu2026:VIA,Jia2026,yang2026astranomy}. We refer to this category as Agentic Manipulation Policies (AMPs).

In this paper, we systematically analyze agentic manipulation policies across models and harness configurations. We present \sysname, a reference harness that provides visual observations, code execution, and direct control of manipulator motion. Paired with frontier models, \sysname achieves state-of-the-art zero-shot results on the RoboLab-120~\cite{robolab} and CaP-Gym~\cite{capx} benchmarks.
The agent succeeds in 104 of 108 real-world trials spanning twelve tasks and three robot instances, compared to 49 for FLUX 3 Action~\cite{flux3a} and 30 for $\pi_{0.5}$~\cite{pi05}, thus reducing the failure rate by 15x and 20x, respectively, relative to these WAM and VLA baselines.
See \cref{fig:first} and the supplementary video for curated highlights.

We document emergent behaviors that arise as the agent brings its broader cognitive faculties to bear on manipulation. When a book lies flat on a table and cannot be grasped as is, the agent pushes it to the table edge to expose a graspable surface. When a target object is hidden, the agent moves occluding objects out of the way to find it. When a large fruit is too wide to grasp, the agent pins it against a heavier object, slides the gripper underneath, and lifts it like a forklift. Failed actions can themselves supply information: the agent infers an unseen shelf from a blocked motion and changes its grasp and approach. On the real robot, the agent diagnoses camera calibration errors from discrepancies between commanded and observed motion, then compensates for them in subsequent actions. None of these behaviors are specified in the prompt or supplied as skills in the harness. The generalist model develops them within each episode by reasoning about geometry, contact, and the consequences of its own actions.

We also investigate few-shot and continual learning from the agent's own experience. We equip \sysname with a persistent memory, allowing the agent to record useful observations, strategies, and code after reflecting on its successes and failures. Experience accumulates in this memory bank, while the model's weights remain fixed. No expert demonstrations or gradient updates are required. We evaluate this approach on the 42 bimanual manipulation tasks of RoboDojo~\cite{robodojo}. Practicing individual tasks raises the success rate on unseen layouts from 26.4\% to 84.1\%, above all entries on the benchmark's leaderboard. We also test whether the agent can accumulate knowledge across tasks via continual learning. In this setting, the agent attempts each task once, in sequence, and updates its memory representation in the course of this continual deployment. A single pass through the tasks more than doubles the success rate, to 57.6\%. Learning transfers to tasks not yet attempted without erasing gains on tasks already encountered. Experience improves efficiency as well: an agent with memory uses about 35\% fewer model calls in comparison to zero-shot deployment.

We conduct controlled experiments to isolate the effects of the model, its reasoning effort, and the harness's perception and motion interfaces. The choice of model has the largest effect. For example, \sysname's success rate on RoboLab-120 is 82.8\% with Opus 5.5, 51.6\% with Haiku 5.5, and 35.2\% with Opus 4.5. More reasoning helps all evaluated models. The value of additional harness support depends on the model. Weaker models are helped by assistive grasping and placing routines, while stronger models perform best without these routines. Stronger models also cope better with missing sensory information. With the \sysname harness, Opus 5.5 achieves a success rate of 82.8\% with full sensory input, 81.1\% without direct depth input, and 75.8\% without depth or camera calibration. While the overall success rate does not suffer much, sensory impairment takes a toll on efficiency. The agent routes around limited sensory input by reconstructing 3D points via triangulation across views and, when needed, estimates camera calibration from images taken at known gripper poses. The mean number of model calls per episode rises from 34.8 with full sensory input to 46.3 without direct depth input and 53.5 without depth or camera calibration. A weaker model cannot compensate for impoverished perception: the success rate of \sysname with Haiku 5.5 on RoboLab-120 collapses from 51.6\% with full sensory input to 10.7\% without depth or camera calibration. Our experiments also indicate that replacing visual input with textual descriptions has a sharply detrimental effect, for example cutting the success rate of \sysname with Opus 5.5 from 82.8\% to 36.8\%; this contrasts with prior work, which recommended replacing visual input with textual descriptions~\cite{capx}.

Our results point to agentic manipulation policies as a promising foundation for generalist manipulation. AMPs are not yet practical, as the robot spends much of each episode waiting for the model to reason. But neither are the alternatives: the strongest VLA and WAM entries on the RoboLab-120 leaderboard at the time of writing succeed in 49.2\% and 42.9\% of the episodes, respectively, in contrast to \sysname's 85.6\%. A policy that fails every other task is not practical, however quickly it does so. We see a more promising path in making capable policies fast. Recent studies estimate that the cost of a given capability level has fallen by roughly an order of magnitude per year~\cite{Gundlach2025,EmbersonRoodman2026}. A harness could further reduce waiting by reusing generated code, running feedback loops locally~\cite{Liang2023,Huang2023:VoxPoser,Huang2026}, and planning the next motion while the current one executes~\cite{Black2025:RTC}. AMP experience could also be used to synthesize closed-loop onboard controllers~\cite{Huang2026} or train compact models through distillation~\cite{Shen2026:EmbodiedSWE,Liu2026:Guava}. Our analysis suggests that agentic manipulation policies are substantially more general and robust than alternative paradigms, opening the way to broad deployment of generalist manipulation capabilities across hardware embodiments and application domains.

\begin{widefigure} %
	\centering
	\includegraphics[width=\linewidth]{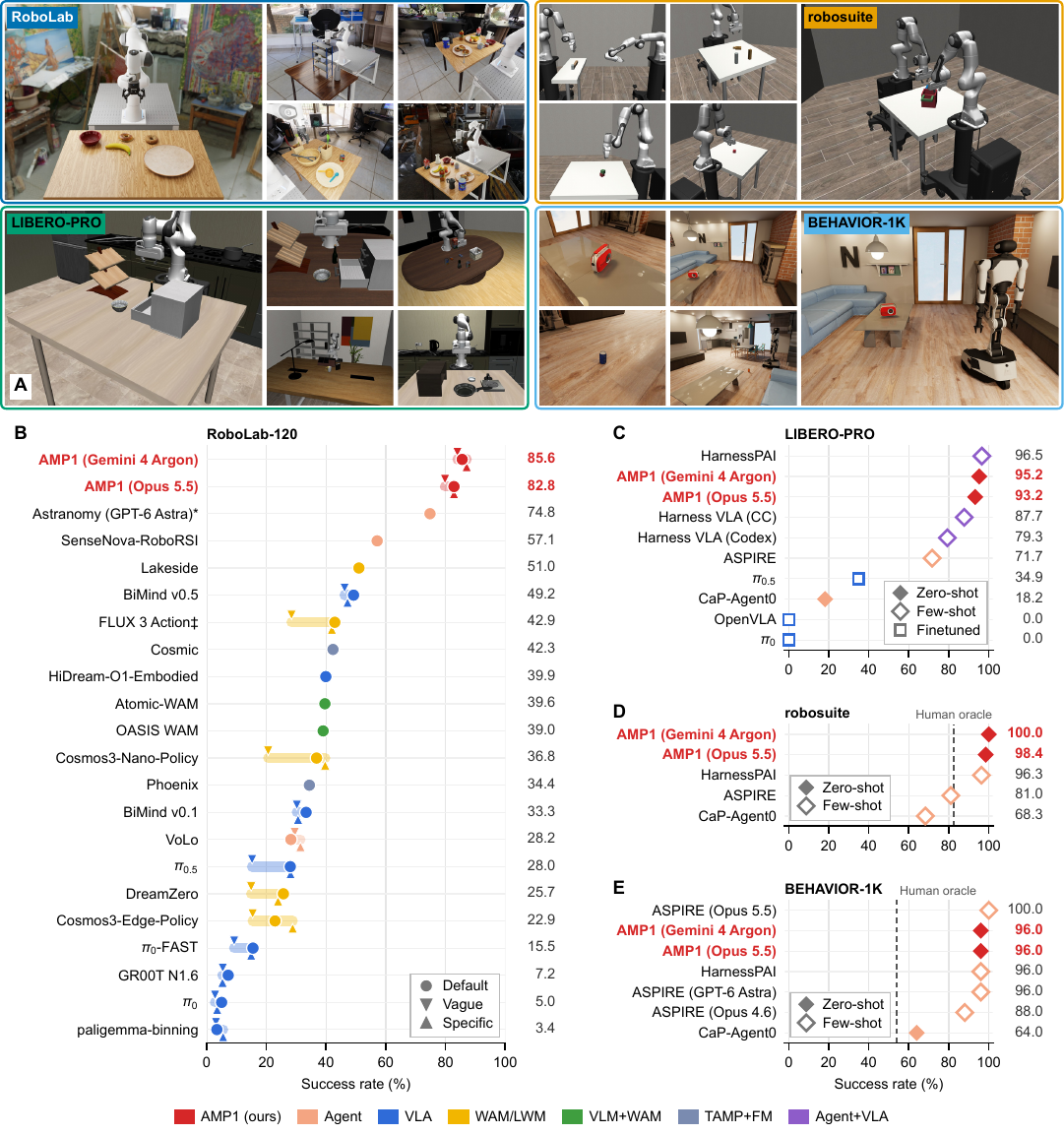}
    \caption{\textbf{Simulation evaluations.}
  		(\textbf{A}) We evaluate \sysname on four open-source simulation benchmarks: RoboLab-120~\cite{robolab} and the LIBERO-PRO, robosuite, and BEHAVIOR-1K suites of CaP-Gym~\cite{capx}. The tasks span short and long horizons, one and two arms, fixed and mobile bases, and prehensile and non-prehensile manipulation.
          (\textbf{B}) Success rates on RoboLab-120 with default ($\bullet$), vague ($\blacktriangledown$), and specific ($\blacktriangle$) instructions. Numbers on the right show the success rate with default instructions. \sysname has the highest success rate on the benchmark and the smallest relative spread across the three instruction conditions among entries that report all three ($*$ not an official submission on the RoboLab-120 leaderboard; $\ddagger$ our evaluation on the vague and specific instruction conditions).
          (\textbf{C} to \textbf{E}) Evaluation on the CaP-Gym suites, zero-shot and with the same harness as on RoboLab-120 ($\blacklozenge$: zero-shot, without training on the benchmark or rollouts of the evaluated tasks before testing; $\lozenge$: few-shot, learning from rollouts of the evaluated tasks; $\square$: fine-tuned on the benchmark's demonstrations; dashed lines: programs written by human experts~\cite{capx}).
          (\textbf{C}) On LIBERO-PRO, \sysname achieves a 95.2\% success rate, outperforming the zero-shot CaP-Agent0 baseline as well as fine-tuned VLAs and all but one few-shot agent baselines.
          (\textbf{D}) On robosuite, which includes both single-arm and bimanual tasks, \sysname achieves a 100.0\% success rate zero-shot, outperforming all baselines.
          (\textbf{E}) On the two BEHAVIOR-1K mobile manipulation tasks in CaP-Gym, \sysname achieves a 96.0\% success rate, outperforming the zero-shot CaP-Agent0 baseline by a wide margin and matching or exceeding all but one few-shot baselines.
      }
	\label{fig:simeval}
\end{widefigure}

\section{Results}
\label{sec:results}

\subsection{\sysname achieves state-of-the-art zero-shot results}
\label{sec:main-results}

\subsubsection{Benchmarks}
\label{sec:results:benchmarks}

We chose four simulation benchmarks that facilitate comprehensive quantitative evaluation of \sysname against the most effective learned policies and other agents (\cref{fig:simeval}A). RoboLab-120
comprises 120 manipulation tasks specified in natural language~\cite{robolab}. The tasks span visual, relational, and procedural competencies; each is posed with default, vague, and specific instructions. The other three benchmarks are suites of CaP-Gym~\cite{capx}: LIBERO-PRO perturbs object positions and instructions to test whether a policy generalizes beyond familiar scenes~\cite{libero_pro}; robosuite has five unimanual and two bimanual tasks~\cite{robosuite};
and two tasks from BEHAVIOR-1K that require mobile manipulation in a simulated home~\cite{li2024behavior1k}. We run 10 episodes per task on RoboLab-120 (3,600 in total across the default, vague, and specific instruction conditions), 50 on LIBERO-PRO (3,000 in total), 100 on robosuite (700), and 25 on BEHAVIOR-1K (50). The \sysname harness is the same on all four benchmarks; only the robot description and camera configuration change. \sysname solves every episode zero-shot: in a single attempt, without demonstrations, training, or practice rollouts.

  \begin{table*}[t]
        \centering
        \small
        \caption{\textbf{Real-world trials.}
        Evaluation on diverse tasks on physical hardware.
        `Average time to success' is the mean over successful trials only, in seconds from the start of the episode to task completion.
        A dash (--) indicates that the method did not succeed on any trial of that task.
        The timing for \sysname is decomposed into execution (Exec.)\ and reasoning (Reason.).
        \sysname is paired with Opus 5.5 for these experiments.
        \sysname is evaluated with robot error recovery enabled; the baseline deployments terminate trials when robot controller errors interrupt motion. An ablation without error recovery is reported in \cref{tab:results:rfe}.
        The legend below the table lists the various task features.}
        \label{tab:results:realworld}
        \setlength{\tabcolsep}{2pt}
        \setlength{\aboverulesep}{0pt}   %
        \setlength{\belowrulesep}{0pt}
        \renewcommand{\arraystretch}{1.3}   %
        \begin{tabular*}{\textwidth}{@{\extracolsep{\fill}}>{\raggedright\arraybackslash}p{\dimexpr\textwidth-312pt\relax}||ccc|ccc|ccccc@{}}
            \toprule
            & \multicolumn{3}{c|}{Task features} & \multicolumn{3}{c|}{Success} & \multicolumn{5}{c}{Average time to success (seconds)} \\
            \cmidrule(lr){2-4}\cmidrule(lr){5-7}\cmidrule(lr){8-12}
            Task Instruction & Manip. & Horizon & Reasoning
                 & $\pi_{0.5}$ & FLUX & \sysname
                 & $\pi_{0.5}$ & FLUX & \multicolumn{3}{c}{\sysname} \\
            \cmidrule(lr){10-12}
            & & & & & & & & & Exec. & Reason. & Total \\
            \midrule
            Put the rubiks cube to the left of the bowl & \tP       & \tSS & \tSR             & 5 / 9 & 9 / 9 & 9 / 9 & 18  & 20 & 22  & 72  & 94  \\
            Put the cube and the banana in the bowl & \tP       & \tMS & --               & 9 / 9 & 9 / 9 & 9 / 9 & 26  & 30 & 44  & 138 & 182 \\
            Put all the red things in the grey bin & \tP       & \tMS & \tSG             & 5 / 9 & 7 / 9 & 9 / 9 & 53  & 39 & 46  & 146 & 192 \\
            Pack canned foods into the bin & \tP       & \tMS & \tSG             & 2 / 9 & 7 / 9 & 9 / 9 & 78  & 40 & 54  & 154 & 208 \\
            Put all the lemons and the orange on the plate & \tP       & \tMS & \tSG             & 0 / 9 & 0 / 9 & 9 / 9 & --  & -- & 75  & 115 & 190 \\
            Put all the tools except the hammer in the right bin & \tP       & \tMS & \tSR\,\tSG\,\tEX & 3 / 9 & 2 / 9 & 9 / 9 & 143 & 79 & 102 & 301 & 403 \\
            Pour the blue cubes onto the plate & \tP       & \tSS & \tSG\,\tSR       & 1 / 9 & 5 / 9 & 7 / 9 & 34  & 43 & 39  & 196 & 235 \\
            Put the bowl on the plate & \tNP\,\tP & \tMS & \tSR\,\tAF       & 2 / 9 & 0 / 9 & 7 / 9 & 92  & -- & 74  & 410 & 484 \\
            Nest the cups inside the largest cup & \tP       & \tMS & \tOR             & 1 / 9 & 3 / 9 & 9 / 9 & 25  & 38 & 74  & 188 & 262 \\
            Pick up the yellow cube & \tNP\,\tP & \tSS & \tSG\,\tAF       & 2 / 9 & 7 / 9 & 9 / 9 & 214 & 96 & 76  & 244 & 320 \\
            Use the spatula to push the four cubes into a straight, aligned row & \tNP\,\tP & \tMS & \tSR             & 0 / 9 & 0 / 9 & 9 / 9 & --  & -- & 126 & 475 & 601 \\
            Cover each cube with a cup, then uncover the yellow cube & \tP       & \tMS & \tME\,\tSG       & 0 / 9 & 0 / 9 & 9 / 9 & --  & -- & 90  & 136 & 226 \\
            \midrule
            Total & & & & 27.8\% & 45.4\% & 96.3\% & 62 & 44 & 69 & 211 & 280 \\
            \bottomrule
          \end{tabular*}

        \vspace{4pt}
        \begin{minipage}{\linewidth}
            \footnotesize\raggedright
            \hangindent=5.8em \hangafter=1 \makebox[5.8em][l]{\textbf{Manip.}}%
            \tP\hspace{0.3em}prehensile \hspace{1em} \tNP\hspace{0.3em}non-prehensile\par
            \hangindent=5.8em \hangafter=1 \makebox[5.8em][l]{\textbf{Horizon}}%
            \tSS\hspace{0.3em}single-step \hspace{1em} \tMS\hspace{0.3em}multi-step ($\geq 2$ subtasks)\par
            \hangindent=5.8em \hangafter=1 \makebox[5.8em][l]{\textbf{Reasoning}}\ifcsname iflatexml\endcsname\ \fi%
            \tSR\hspace{0.3em}spatial reasoning (a relation or a left/right choice) \hspace{1em}
            \tSG\hspace{0.3em}semantic grounding (color or category) \hspace{1em} \tEX\hspace{0.3em}exclusion \hspace{1em}
            \tOR\hspace{0.3em}ordering (compare sizes, decide an order) \hspace{1em}
            \tAF\hspace{0.3em}affordance (an enabling action is needed before a grasp is possible)
            \hspace{1em}
            \tME\hspace{0.3em}memory (information that is no longer visible)\par
        \end{minipage}
    \end{table*}

On RoboLab-120, \sysname paired with Gemini 4 Argon succeeds in 85.6\% of episodes (95\% confidence interval, 83.5 to 87.5\%), and in 82.8\% when paired with Opus 5.5 (95\% CI, 80.6 to 84.9\%)~-- the highest success rates reported on the benchmark (\cref{fig:simeval}B). The strongest entry on the RoboLab-120 public leaderboard at the time of writing is also an agent: SenseNova-RoboRSI, which is based on GPT-6 Astra~\cite{sensenova}; it has a 57.1\% success rate. The strongest learned policy on the leaderboard, Lakeside, has a 51.0\% success rate, and the strongest VLA and WAM entries are at 49.2\% and 42.9\%, respectively. Astranomy, an agentic manipulation policy evaluated outside the leaderboard, reaches 74.8\% with GPT-6 Astra~\cite{yang2026astranomy}.
Astranomy also reported a 25.3\% success rate with Opus 5.5 on a 30-task subset of RoboLab; this contrasts with the 82.8\% success rate achieved by \sysname with Opus 5.5 on the full RoboLab-120.

\sysname is robust across instruction conditions. In RoboLab's `vague' and `specific' instruction conditions, \sysname succeeds in 84.0\% and 87.1\% of episodes with Gemini 4 Argon, and in 79.8\% and 82.8\% of episodes with Opus 5.5, respectively. The success rate thus varies by about three points across the three conditions (Supplementary Section~\ref{sec:appendix:instruction_spread} and \cref{fig:instruction_spread}). Specialist policies are far more brittle: with vague instructions, FLUX 3 Action~\cite{flux3a} falls from 42.9\% to 28.4\% and $\pi_{0.5}$~\cite{pi05} from 28.0\% to 15.2\%. (Since the RoboLab-120 leaderboard does not report FLUX 3 Action performance under the `vague' and `specific' conditions, we evaluated it ourselves on these conditions using its publicly released checkpoint.)

\sysname's success rate on RoboLab-120 approaches saturation, with the remaining failures bounded largely by benchmark artifacts rather than policy errors (Supplementary Materials, \cref{sec:appendix:robolab_failures}, and \cref{fig:failure_sankey}). Across the 1,200 trials in the `default' instruction condition, 121 of the 173 unsuccessful episodes with Gemini 4 Argon and 135 of the 206 unsuccessful episodes with Opus 5.5 can be traced to issues with the benchmark setup: either the object's geometry or mass defeats a standard parallel-jaw grasp (77 and 101 episodes, respectively) or a flaky verifier rejects an end state that satisfies the instruction (44 and 34 episodes). Only two tasks fail all 10 trials for both models; the remaining affected tasks still succeed in about half of their trials when a placement happens to satisfy the verifier's geometric check or when \sysname improvises a non-standard grasp (\cref{sec:results:emergent_behaviors}). Across the 95 tasks that do not suffer from these benchmark artifacts, \sysname with Gemini 4 Argon succeeds in 96.2\% of episodes (914 of 950).

On CaP-Gym, the strongest agent baselines~-- HarnessPAI~\cite{harnesspai}, ASPIRE~\cite{aspire}, and Harness VLA~\cite{harness_vla}~-- all involve improving a program, a skill library, or a memory representation over rollouts of the evaluated tasks and are thus not zero-shot. In contrast, \sysname solves every episode zero-shot, using the same harness in CaP-Gym as in RoboLab (\cref{fig:simeval}C--E).

On the LIBERO-PRO subset of CaP-Gym, \sysname with Gemini 4 Argon succeeds in 95.2\% of episodes, against 18.2\% for CaP-Agent0~\cite{capx}, the only other zero-shot agent, and 34.9\% for $\pi_{0.5}$, a fine-tuned VLA~\cite{pi05}. Only HarnessPAI~\cite{harnesspai}, which evolves its program on seeds that lie inside the evaluation, scores higher (96.5\%). Its lead comes entirely from LIBERO-Goal (94.6\% and 97.8\% against 81.4\% and 91.4\% under swap and task perturbations, respectively), while \sysname is ahead on LIBERO-Object and LIBERO-Spatial. In almost all of the unsuccessful LIBERO-Goal episodes, \sysname ended the episode believing it had succeeded. The instructions do not say what LIBERO checks, such as which side of the stove counts as its front or how far a drawer must open, and a human review of the episodes marked as failures suggests that \sysname interpreted the instructions in a way that is consistent with human interpretation but inconsistent with the verifier implemented by the benchmark.
HarnessPAI utilizes a VLA that was fine-tuned on LIBERO demonstrations that succeed according to the implemented verifier, thus leaking the verifier's implementation
into the training data.

On the robosuite subset of CaP-Gym, \sysname paired with Gemini 4 Argon succeeds in all 700 episodes, including the two tasks that require coordinating two arms, against 96.3\% for HarnessPAI~\cite{harnesspai} and 82.6\% for programs written and debugged by human experts~\cite{capx}. 
On the two BEHAVIOR-1K tasks in CaP-Gym, \sysname succeeds in 48 of 50 episodes (96\%), against 64\% for CaP-Agent0 and 54\% for human-written programs~\cite{capx}; only ASPIRE~\cite{aspire}, which evolves its programs over rollouts of the same task (and is thus not zero-shot), succeeds in every episode. 

\subsubsection{Real-world trials}

\sysname's strong performance on benchmarks is matched in zero-shot trials on real-world hardware. We provide quantitative and qualitative results from diverse real-world manipulation tasks.   All real-world experiments were conducted using a 7-DoF Franka Research 3 arm equipped with a Robotiq 2F-85 gripper and a three-camera vision system similar to DROID \cite{droid} (left and right shoulder views, plus a wrist-mounted camera). The \sysname harness was paired with Claude Opus~5.5 at high reasoning effort. The results are reported in Table~\ref{tab:results:realworld}.

We designed 12 challenging short- and long-horizon manipulation tasks and evaluated \sysname against leading VLA and WAM policies, $\pi_{0.5}$ and FLUX 3 Action~\cite{pi05,flux3a}. The tasks cover a broad range of capabilities, including prehensile and non-prehensile manipulation, spatial and temporal reasoning, semantic grounding, and object exclusion. We include both single- and multi-step tasks, the latter containing two or more subtasks. The results of the real-world trials are presented in Table~\ref{tab:results:realworld}.
Further details of our experimental setup and evaluation protocols are provided in~\cref{sec:material:protocol:real_robot}. 
All tasks are illustrated in the Supplementary Materials (\cref{sec:appendix:fig:tasks}).

\sysname matches or outperforms the baselines on all 12 tasks. It particularly excels in scenarios that require object identification in crowded settings or complex manipulation strategies, such as non-prehensile interaction. Overall, \sysname achieves a 96.3\% success rate, compared to 45.4\% for FLUX and 27.8\% for $\pi_{0.5}$. Baseline methods struggle to accurately identify target objects and frequently grasp incorrect objects, including those explicitly excluded by the task prompt. For example, when instructed to ``Put all the tools except the hammer in the right bin'', both baselines still occasionally grasp the excluded object.

Most real-world struggles of \sysname are caused by collision, mainly with the surface of the table, the walls of the bin, and self-collision. These collisions occur because the underlying motion planning is sensitive to camera calibration and relies on approximations including simplified robot geometry and unmodeled hardware such as camera mounts and cables.
Collisions can trigger the Franka controller's collision reflex, which interrupts motion when estimated external forces or torques exceed configured thresholds. The resulting error must be cleared before commanded motion can resume. We provide SDK functions through which the agent can query the arm's state and request error recovery. Following successful recovery, the agent can replan and continue the task. With this functionality, \sysname fails in only 4 out of 108 trials. (In two, some cubes fell outside the target plate during pouring. In the other two, a bowl rolled off the table.) Without these functions, our evaluation protocol terminates a trial when a robot controller error interrupts motion; in this condition, \sysname succeeds in 94 of 108 trials (87.0\%; \cref{tab:results:rfe}).

A primary advantage of \sysname is its capacity for dynamic error recovery and replanning as the task unfolds. For instance, for a task that requires moving a flipped bowl placed on top of a flat bin, where no initial feasible grasp exists, \sysname recognizes that a direct grasp is infeasible and pushes the bowl to hang from the side of the bin where it can be grasped, thus combining non-prehensile and prehensile techniques (\cref{fig:emergent_behavior_appendix_1}A). In contrast, baseline methods often continue pursuing hard or infeasible strategies (e.g., an intractable grasp) until the episode times out. Furthermore, for tasks that require complex semantic understanding, such as covering and uncovering multiple cubes with cups, the baseline methods perform inconsistent actions indicative of challenges with task comprehension. $\pi_{0.5}$ is particularly prone to exhibiting indecisive behavior, repeatedly grasping and dropping a correctly identified target object without completing the placement. 

A number of case studies, documented in~\cref{fig:first} and the supplementary video, demonstrate that \sysname can generalize zero-shot to even more complex tasks that require sequential dependency planning via interactive perception: manipulating the environment to uncover and subsequently manipulate hidden objects (\cref{fig:first}A), devising effective tool-use strategies (\cref{fig:first}B), reasoning through and physically executing a winning strategy in a game of chess followed by putting away the pieces (Fig. \ref{fig:first}C), pushing or sliding to reveal occluded grasps (\cref{fig:first}D), and manipulating articulated objects (\cref{fig:first}E).

\subsection{Emergent behaviors}
\label{sec:results:emergent_behaviors}

Beyond aggregate task performance, we examine the manipulation strategies developed by \sysname during execution. None of the behaviors described below was specified in the prompt or provided as a predefined skill in the harness. The agent develops these strategies while solving tasks, combining geometric reasoning, visual feedback, and physical interaction to construct higher-level manipulation skills. We highlight several examples, some of which are shown in \cref{fig:emergent_behavior} and the supplementary video.

    \mypara{Searching for hidden objects} When the instructed object is not visible from any camera, \sysname does not stop at what it sees. It reasons about which objects are large enough to hide the target, moves them aside one at a time, and looks again after each move until the target appears (\cref{fig:emergent_behavior}A). The robotics literature calls this mechanical search~\cite{danielczuk2019mechanical}, a form of interactive perception~\cite{Bohg2017}: the agent acts to gain information, not only to make progress on the task.

    \mypara{Moving the goal instead of the object} When an object cannot be grasped, \sysname sometimes brings the goal to the object instead. In one task, ``Put the bowl on the plate'', the bowl lay upside down on a bin, too wide for the gripper and with no edge the fingers could reach under (\cref{fig:emergent_behavior}B). \sysname instead grasped the plate by its rim and slid it up against the side of the bin, then pushed the bowl across the bin with the open gripper until it fell off the edge onto the plate. The bowl was never held: it reached its goal by non-prehensile manipulation alone.

\begin{widefigure} %
	\centering
    	\includegraphics[width=\linewidth, trim=0pt 0pt 0pt 0pt, clip]{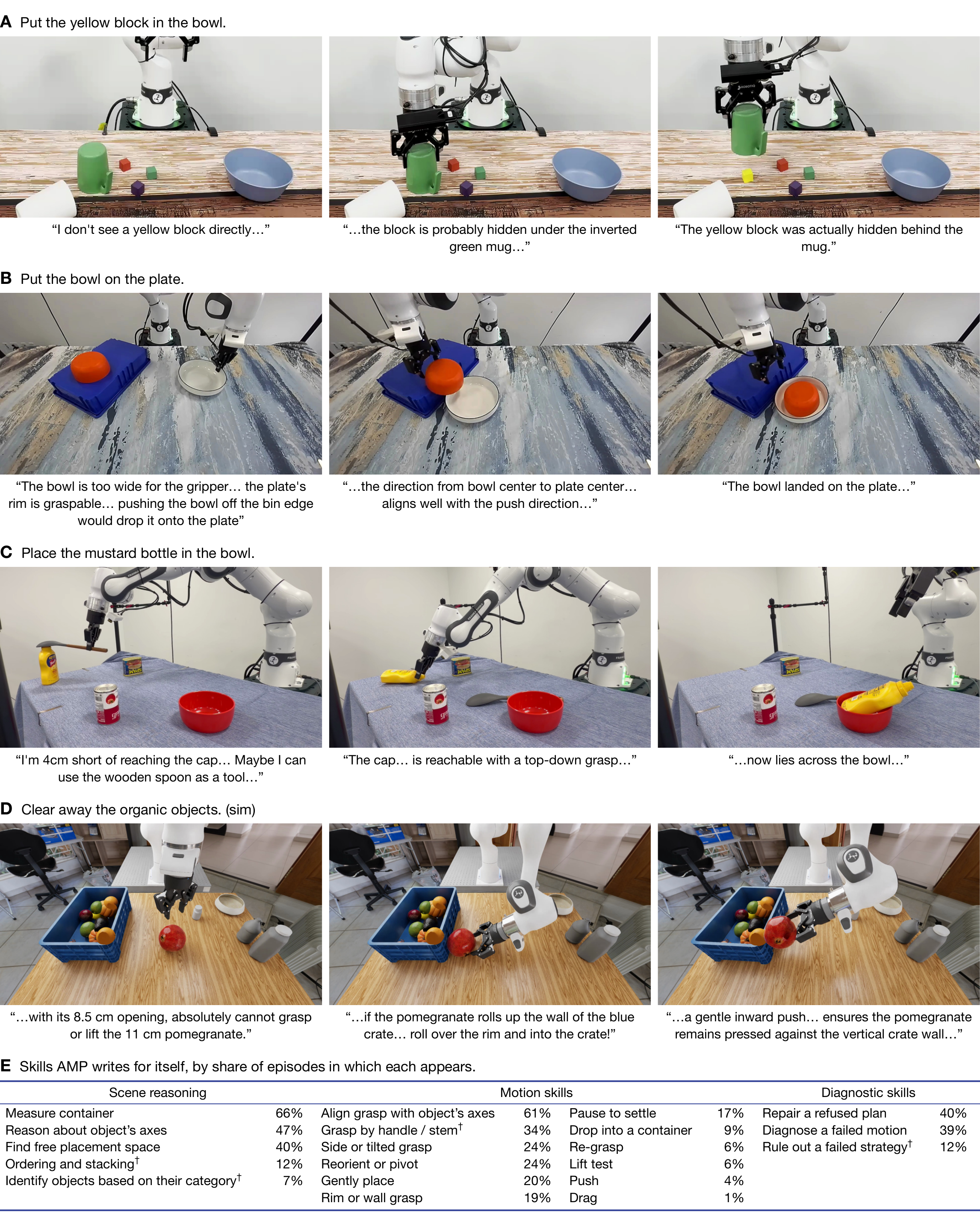}

	\caption{\textbf{Emergent behaviors in \sysname.} Four episodes in which \sysname overcomes an unforeseen challenge: (\textbf{A} to \textbf{C}) on the real robot and (\textbf{D}) in RoboLab simulation. Quotes under each image are from the agent's thinking traces. (\textbf{A}) The target is hidden from view, so \sysname reasons about where it could be hidden and moves the occluding object to look. (\textbf{B}) The bowl offers no viable grasp, so \sysname moves the plate against the bin instead and pushes the bowl off the higher bin edge onto the plate. (\textbf{C}) The bottle is out of the arm's reach, so \sysname uses a wooden spoon to tip it over toward itself, then grasps it by the cap. (\textbf{D}) The pomegranate is too wide to grasp, so \sysname presses it against the wall of the crate and rolls it up the wall. (\textbf{E}) Skills that \sysname writes for itself, anew in each episode, and the share of the 1,200 RoboLab episodes in which each appears. $^{\dagger}$Appears only in its reasoning, not as code.}
	\label{fig:emergent_behavior}
\end{widefigure}

    \mypara{Using objects as tools} In some cases in which an object lies beyond the arm's reach, \sysname can use another item in the scene as a tool. It first establishes from inverse kinematics that the object is out of reach, and by how much. It then searches the scene for a tool long enough to cover that distance and reasons about how to use it to apply force, for example pulling rather than pushing so that the object moves toward the arm (\cref{fig:emergent_behavior}C). Once the object is within reach, it sets the tool aside and grasps the object directly. None of this is described in the prompt, suggesting that \sysname can at times improvise tool use from its understanding of geometry and contact. This strategy parallels tool selection in New Caledonian crows, which choose sticks long enough to retrieve otherwise inaccessible food~\cite{ChappellKacelnik2002}, and has precedents in robotics~\cite{Xu2023}.

    \mypara{Scooping objects too wide to grasp} When an object, such as a large pomegranate, is wider than the gripper's 85-mm stroke in every direction, \sysname sometimes pins it against a heavier object so that it cannot roll or slide away (\cref{fig:emergent_behavior}D). It then slides the open gripper beneath it and lifts it like a forklift to carry it to the goal. Like pushing an object to the table edge, this strategy exploits contact with the environment to achieve a grasp the gripper cannot achieve alone, a capability known as extrinsic dexterity~\cite{dafle2014extrinsic}.

    \mypara{Pushing to enable grasping} When an object is too wide to grasp where it lies, \sysname pushes it toward the edge of its supporting surface until part of it overhangs, giving the gripper a narrower part to hold (\cref{fig:emergent_behavior_appendix_1}A). When objects stand too close together to grasp one at a time, it pushes them apart (\cref{fig:emergent_behavior_appendix_1}B). In both cases, \sysname treats the scene as something it can rearrange rather than a fixed input: when no grasp is feasible, it first acts to create one, a strategy known as pre-grasp manipulation~\cite{king2013pregrasp}.

    \mypara{Correcting calibration errors} On the physical robot, camera calibration is imperfect, and the resulting errors in estimated poses can be too large for precise manipulation. When a motion misses its target, \sysname compares where the gripper appears in the images with where it was commanded to go, attributes the offset to the calibration and corrects it (\cref{fig:emergent_behavior_appendix_1}C). It then uses the corrected calibration for the rest of the episode. The same reasoning lets it calibrate the wrist camera from scratch when no calibration is given (Supplementary Materials, \cref{sec:appendix:case_studies:self_calibration}).   

    \mypara{Gentle placement of long objects} When placing a long object, \sysname first rests one end on the surface, then lowers the other along an arc (\cref{fig:emergent_behavior_appendix_1}D). This strategy uses surface contact to support the object during lowering, suggesting reasoning about contact and the physical consequences of the placement motion.

    \mypara{Reorienting objects by hooking and pivoting the gripper} For objects lying on their side, \sysname sometimes closes the gripper, inserts the closed fingers into the object's opening, and uses them as a hook while moving the gripper along an arc until the object stands upright (\cref{fig:emergent_behavior_appendix_2}A). For objects placed upside down, it first grasps an outer feature to tip the object onto its side, then hooks and pivots it. 

    \mypara{Inferring environmental constraints from physical interaction} In one task, \sysname must place an object on a shelf whose top is not visible from any camera, so that part of the shelf is missing from its point cloud. Its first placement motion fails when the arm is blocked by this unseen geometry. From the gap between the commanded and reached end-effector poses, \sysname infers that an upper shelf lies just outside its field of view, assumes it sits at the edge of the visible region to leave a safety margin, and adjusts its grasp and approach angle accordingly (\cref{fig:emergent_behavior_appendix_2}B). \sysname thus treats a failed action as a measurement, using physical interaction to recover information its cameras cannot provide. Together with its behavior in the wrist-camera setting (\cref{sec:results:analysis:perception}), where it moves the arm back to see more of the scene, this suggests that \sysname uses action as a source of information, not only as a means of changing the scene.

    \mypara{Mechanical interlocking grasps for heavy objects} Through interaction with the environment, \sysname discovers that friction-based grasps are brittle for heavy objects and prone to slippage. When the object's geometry allows it, \sysname switches to a mechanical interlocking grasp. For example, to lift a heavy drill, instead of grasping it from the top, it closes the fingers around the top of the handle, just below the drill's head (\cref{fig:emergent_behavior_appendix_2}C). Upon lifting, the head rests on the fingers, which yields an inherently stable grip that does not depend on friction.

Beyond these examples, we measured how often \sysname composes the same building blocks across many tasks (\cref{fig:emergent_behavior}E). We matched the code \sysname executed and the intentions stated in its reasoning, in 1,200 RoboLab episodes, against a set of skill patterns, and report a pattern as a skill only if it recurs (a) in at least five episodes and (b) in at least two tasks; the first condition excludes rare improvisations and spurious matches, and the second excludes procedures that serve a single task. Because the harness provides no grasping, placing, or pushing primitives (\cref{sec:material:motion}), \sysname builds these skills itself, anew in each episode (\cref{fig:emergent_behavior}E). (These statistics are reported for \sysname paired with Opus~5.5.) The most frequent skills are routine geometric habits: \sysname measures a container before placing into it in 66\% of episodes, aligns its grasp with the object's axes in 61\%, and reasons about an object's long and short axes in 47\%. When the harness rejects a proposed motion, \sysname reads the reason and repairs the plan (40\% of episodes). When a motion fails during execution, it diagnoses the cause before replanning (39\%).

Together, these observations suggest that \sysname can synthesize manipulation skills from the perception and control interfaces it is given, adapting its strategies to object geometry, execution failures, and contact constraints. Since these behaviors were neither specified in the prompt nor supplied as predefined skills, they provide qualitative evidence that code generation enables the agent to translate its reasoning into task-specific manipulation procedures. Many such skills recur across RoboLab-120 episodes (\cref{tab:skills}).

\subsection{Learning from Experience}
\label{sec:results:practice}
Thus far, the agent has been evaluated zero-shot, starting every episode from scratch with no memory of its past experience. In the field, however, a robot may be able to build on its experience. We therefore investigate whether the agent can improve with experience, both within a task via few-shot learning and across tasks via continual learning. In the few-shot setting, the agent practices on the same task with resets. In the continual setting, it learns across different tasks as it tackles them in sequence.

We evaluate learning from experience on RoboDojo~\cite{robodojo}, a challenging benchmark that comprises 42 simulated bimanual manipulation tasks across five capability dimensions (\cref{sec:material:protocol:robodojo}).
RoboDojo provides many layouts per task, so a given task can be practiced on certain scenes and tested on others. Unlike in RoboLab-120, the agent receives only RGB images, without depth or camera calibration. While \sysname approaches saturation on RoboLab-120 zero-shot (\cref{sec:results:benchmarks}), RoboDojo leaves ample headroom for learning from experience.

In both the few-shot and the continual settings, the agent keeps whatever it finds useful in a persistent memory $\mathcal{M}$, a folder that it reads at the start of every episode. We impose no format on $\mathcal{M}$, so it may hold notes, code, or data in any form. Typically, the agent decides to keep a Markdown file of notes (strategies that succeeded or failed, pitfalls and observations, useful constants and parameters) and a Python library of code it found useful. Later episodes can read the notes and import and call the Python code. We refer to the harness augmented by the persistent memory $\mathcal{M}$ as \sysnameexp. In the experiments, \sysnameexp runs Claude Opus~5.5 at high reasoning effort.

In the few-shot setting, we let the agent learn a task by practicing it on several layouts until the agent judges that it can reliably succeed or reaches its practice budget. Each practice episode presents a new layout. The only feedback the agent receives is the benchmark's official binary reward for each episode.
This requires no expert demonstrations, no additional model training, and no reward shaping.

In the continual setting, we test whether the agent can accumulate useful knowledge over its lifetime, applying lessons from earlier tasks to later ones. In this setting, the agent tackles an episode of each of the 42 RoboDojo tasks in random order. A persistent memory store $\mathcal{M}$ is carried across tasks and the agent is instructed to keep only general lessons that may help with any task in the given robot embodiment.

Our approach to learning from experience is gradient-free: the agent revises its memory and code, while the model weights remain fixed. This differs structurally from gradient-based approaches to few-shot learning~\cite{Finn2017} and continual learning~\cite{Kirkpatrick2017,LopezPaz2017}, which adapt neural network weights to new experience. Learning without weight updates has precedents in agents that retain verbal reflections~\cite{Shinn2023,Zhao2024} or accumulate executable skills~\cite{Wang2024:Voyager}. In robotics, ASPIRE retains validated repair knowledge~\cite{aspire}, RoboSkill consolidates execution records into reusable guidance and code~\cite{Xie2026}, and RAX builds autonomous control programs and a persistent skill library through exploration~\cite{Huang2026}. ENPIRE automates policy improvement on physical robots, including both program refinement and gradient-based policy training~\cite{Xiao2026}. We study gradient-free few-shot and continual learning with the generalist model continuing to reason and command the manipulator during execution, measuring both improvement on practiced tasks and transfer and retention across tasks.

\begin{figure*}[tp]
  \centering
  \newcommand{\figletter}[3]{\put(#1,\lxtop{#2}){\makebox(0,0)[tl]{\setlength{\fboxsep}{1pt}\colorbox{white}{\sffamily\bfseries\fontsize{9}{9}\selectfont #3}}}}%
  \setlength{\unitlength}{\linewidth}%
  \begin{picture}(1,0.09743)
    \put(0,0){\includegraphics[width=\linewidth]{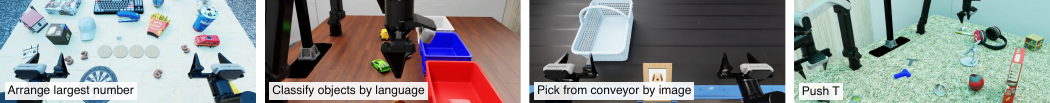}}
    \figletter{0}{0.09743}{A}\put(0.024,\lxtop{0.09743}){\makebox(0,0)[tl]{\setlength{\fboxsep}{1pt}\colorbox{white}{\sffamily\fontsize{7.3684}{8.5}\selectfont RoboDojo}}}
  \end{picture}\\[1pt]
  \begin{picture}(1,0.20621)
    \put(0,0){\includegraphics[width=\linewidth]{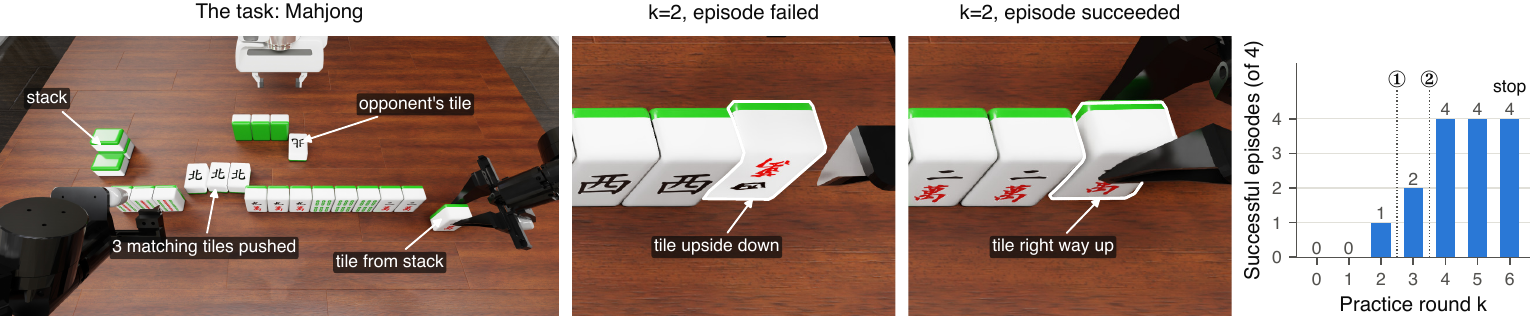}}
    \figletter{0}{0.20621}{B}\figletter{0.3739}{0.20621}{C}\figletter{0.5937}{0.20621}{D}\figletter{0.8135}{0.20621}{E}
  \end{picture}\\[1pt]
  \begin{picture}(1,0.25490)
    \put(0,0){\includegraphics[width=\linewidth]{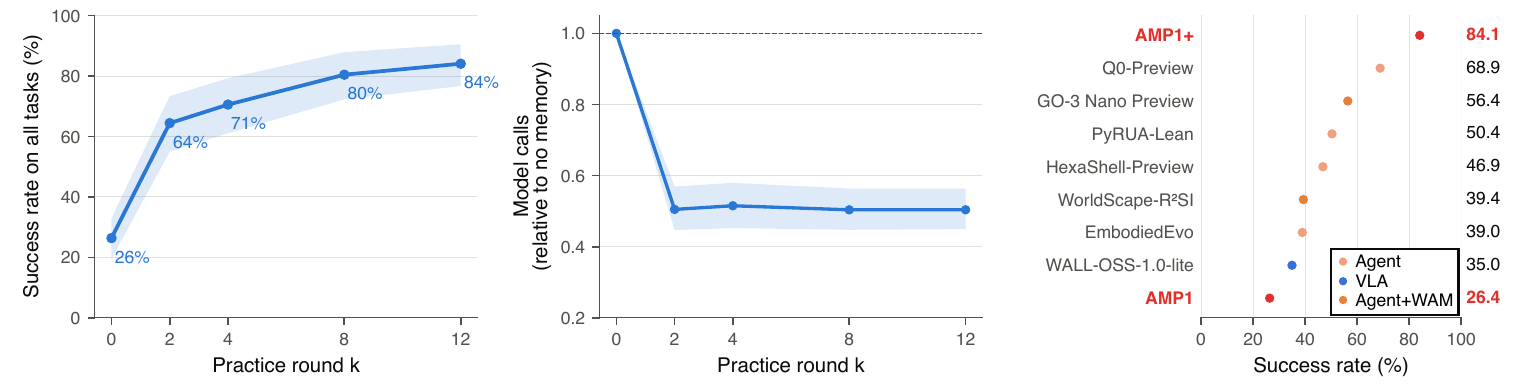}}
    \figletter{0}{0.25490}{F}\figletter{0.326}{0.25490}{G}\figletter{0.652}{0.25490}{H}
  \end{picture}\\[1pt]
  \begin{picture}(1,0.25490)
    \put(0,0){\includegraphics[width=\linewidth]{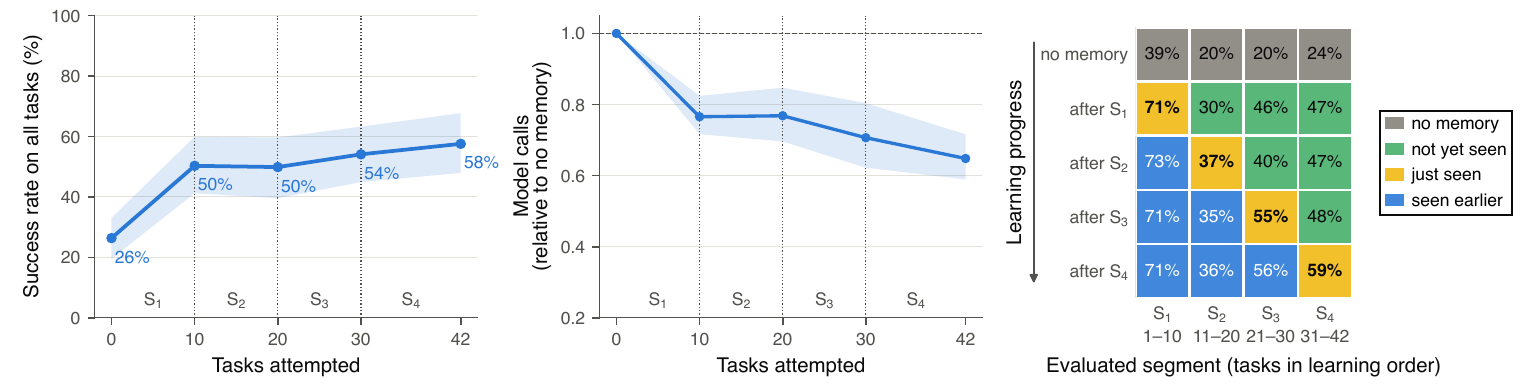}}
    \figletter{0}{0.25490}{I}\figletter{0.326}{0.25490}{J}\figletter{0.652}{0.25490}{K}
  \end{picture}\\[1pt]%
  \begin{picture}(1,0.17157)
    \put(0,0){\includegraphics[width=\linewidth]{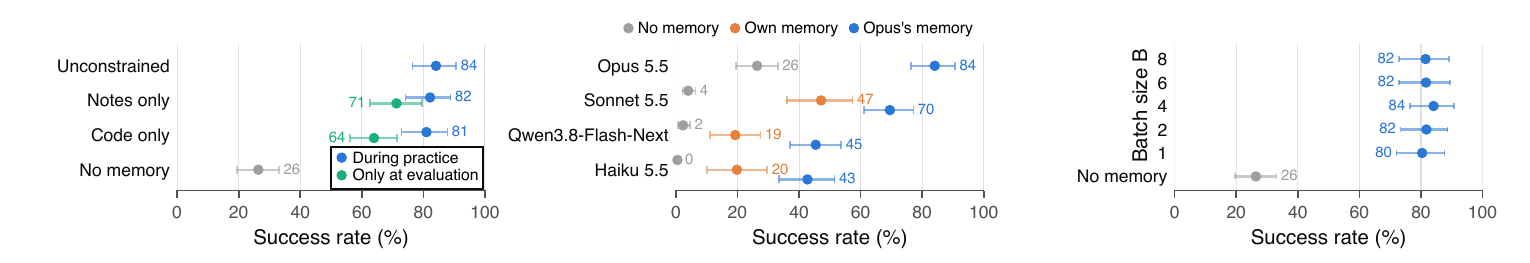}}
    \figletter{0}{0.17157}{L}\figletter{0.326}{0.17157}{M}\figletter{0.652}{0.17157}{N}
  \end{picture}%
    \caption{\textbf{Learning from experience.}
    \textbf{(A)} Example RoboDojo simulation tasks.
    (\textbf{B} to \textbf{E}) Few-shot learning on the Mahjong task. \textbf{(B)} \sysnameexp must push over the three tiles that match the opponent's tile, take a tile from the stack and stand it at the right end of the row. \textbf{(C, D)} Comparing two episodes from the same round ($k=2$); from the binary success signal (one episode failed, the other succeeded), the agent learns that it must stand the tile the right way up.  \textbf{(E)} Successful practice episodes per round. \sysnameexp records the right-way-up rule after round 2 \ding{172} and a rule about when to open the gripper after round 3 \ding{173}, and stops practicing after round 6.
    (\textbf{F} to \textbf{H}) Few-shot learning results. After practice, \sysnameexp reaches a success rate of 84.1\%, up from 26.4\% zero-shot. Practice is run five times per task, and the run that succeeds most often is evaluated. \textbf{(F)} Success rate after $k$ rounds of practice. \textbf{(G)} Number of model calls on successful episodes, relative to zero-shot deployment. \textbf{(H)} Success rate of \sysnameexp and the top entries on the RoboDojo simulation leaderboard (as of November 6, 2026).
    (\textbf{I} to \textbf{K}) Continual learning across tasks, from a single run over one random permutation of the tasks. \textbf{(I)} Success rate through continual learning. \textbf{(J)} Number of model calls relative to zero-shot deployment. \textbf{(K)} Per-segment success rates at the end of each segment, colored by whether the segment has not yet been seen (green), was just seen (yellow), or was seen earlier (blue).
    (\textbf{L} to \textbf{N}) Variations of few-shot learning, each with five practice runs per task and the run selected as in F. \textbf{(L)} Memory restricted to code only or notes only, during practice or only at evaluation. \textbf{(M)} Sonnet~5.5, Qwen3.8-Flash-Next and Haiku~5.5 with no memory, with memory from their own practice, and with the memory written by Opus~5.5. \textbf{(N)} Batch size $B$.
    In \textbf{(F, G, I, J)}, shaded bands, and in (\textbf{L} to \textbf{N}), error bars show 95\% confidence intervals across tasks. All evaluations are on unseen layouts.}
  \label{fig:practice_all}
  \label{fig:practice}
  \label{tab:practice}%
  \label{fig:kong}
  \label{fig:continual-learning}
\end{figure*}

\mypara{Few-shot learning}
\sysnameexp practices each task over several rounds. To save wall-clock time, each round runs a batch of $B$ episodes in parallel. We build the memory $\mathcal{M}_k$ iteratively over rounds ${k = 0, 1, \ldots}$, starting from an empty memory store $\mathcal{M}_0$ (zero-shot). At round $k$, $B$ instances of \sysnameexp, each with memory bank $\mathcal{M}_k$, run on different practice layouts, and the benchmark scores each episode with a binary reward $r_{k,b} \in \{0, 1\}$, $b = 1, \dots, B$. We denote by $\tau_{k,b}$ the episode's transcript, the turn-by-turn record of its messages, code, outputs, and camera images. When the round ends, one of the $B$ \sysnameexp instances from the batch continues after its episode to reflect on the round, with its own episode still in context. It sees all rewards and transcripts so far ($r_{j,b}$ and $\tau_{j,b}$ for $0 \le j \le k$ and $1 \le b \le B$), and hence the success rate of each round, as well as the memories $\mathcal{M}_0, \dots, \mathcal{M}_k$. It then populates $\mathcal{M}_{k+1}$ and decides whether to stop practicing. Practice ends when the reflecting agent decides to stop or after $N/B$ rounds, for a budget of $N$ practice episodes. In our experiments, we use $B=4$ and $N=48$, so practice takes place over up to 12 rounds. The memory bank produced by the ultimate practice round is then frozen as $\mathcal{M}^\star$ and used for evaluation. \sysnameexp itself, in practice and at evaluation, sees only its memory, $\mathcal{M}_k$ or $\mathcal{M}^\star$. The rewards and transcripts are available only to the reflecting agent. 

Practice raises \sysnameexp's success rate on RoboDojo from 26.4\% zero-shot to 84.1\%, above every entry on the RoboDojo leaderboard. The strongest leaderboard entry at the time of writing, Q0-Preview, reaches 68.9\%.
\cref{fig:practice}F and G show how success rates and efficiency change over the rounds of practice.
We measure efficiency by the number of model calls per episode. Each call is one request from the harness to the generalist model to provide its next response. The harness then executes any tool calls specified in that response, such as running code that reads camera images, queries the robot's pose, computes trajectories, or moves the robot.
Efficiency improves almost entirely in the first rounds. By $k=2$, the memories of 39 of the 42 tasks hold a code library, and later episodes call it instead of rediscovering the procedure, which halves the number of model calls. Later rounds fix remaining failures and continue to raise the success rate without further reducing the number of model calls per episode. Practice ends early for 26 of the 42 tasks, and the other 16 use the full budget of practice rounds.

\cref{fig:kong}B--E walk through one example, a task drawn from Mahjong. The agent is asked to take a tile from the stack and stand it upright at the right-hand end of its row; it is not told that the tile must also be the right way up, which the task's scoring requires. At $k=2$, one practice episode fails and another succeeds. In its reflection, the agent enlarges the last camera frame of each and sees that the tile's red symbol is at the bottom in the successful episode, like its neighbors', but on top in the unsuccessful episode. Tracing both episodes' actions, it finds that they picked up the face-down tile from opposite ends, and records which end is the tile's top. After one more round, in which the agent learns to open the gripper before reaching for the tile on the stack, every practice episode succeeds, and success on held-out layouts rises from 10\% prior to practice to 100\%.

\cref{fig:practice}L--N present controlled experiments, each with five practice runs per task and the run selected as in \cref{fig:practice}F. Restricting the memory to code or to notes during practice barely changes the success rate (81.0\% and 82.2\%, against 84.1\%), because \sysnameexp adapts to record its experience in the specified medium, keeping notes as code comments or describing procedures precisely enough to rewrite the code. Introducing the restriction post-hoc at evaluation time lowers the success rate to 64.0\% and 71.2\%, respectively (\cref{fig:practice}L). Weaker models gain less from their own practice, reaching 47.2\% (Sonnet~5.5), 19.3\% (Qwen3.8-Flash-Next), and 19.8\% (Haiku~5.5); they are able to take advantage of memory banks produced by Opus~5.5, however, with success rates at 69.5\%, 45.4\%, and 42.7\%, respectively (\cref{fig:practice}M). The batch size barely matters, as every value of $B$ we tested, from 1 to 8, with the same budget of 48 episodes, reaches a success rate between 80.4\% and 84.1\% (\cref{fig:practice}N).

\mypara{Continual learning}
Starting from an empty initialization $\mathcal{M}_0$, \sysnameexp attempts the $t$-th task ($t = 1, \ldots, 42$) with memory bank $\mathcal{M}_{t-1}$, is told whether it succeeded, and produces an updated memory bank $\mathcal{M}_t$.
After one pass through a random permutation of the 42 tasks, the agent's success rate on all tasks rises from 26.4\% zero-shot to 57.6\% with $\mathcal{M}_{42}$.
To track the gain over the sequence, both on tasks that have already been seen and on tasks that have not yet been seen, we split the task sequence into four segments: $S_1$ (tasks 1--10), $S_2$ (11--20), $S_3$ (21--30), and $S_4$ (31--42). At the end of segment $i$, we evaluate \sysnameexp with the memory bank $\mathcal{M}_{10i}$ on all 42 tasks. Most of the gain in the success rate arrives with the first segment. Experience helps with efficiency as well (\cref{fig:continual-learning}I, J).

\Cref{fig:continual-learning}K breaks this gain down segment by segment. Segments that have not yet been seen benefit along with segments that the agent already worked through.
This implies that much of what the agent learns from earlier tasks transfers to new tasks. For example, success rate on $S_4$ rises from 24\% without memory to 47\% after performing $S_1$ alone. Segments seen earlier (blue) stay well above their no-memory success rates, so learning on later segments does not erase earlier gains.

What \sysnameexp writes into its memory explains these results. From the earliest episodes, it records embodiment-specific knowledge, such as the robot's coordinate frames, its physical limits, and how to work around the lack of depth sensing and camera calibration. It also writes code for basic manipulation primitives, such as grasping and placing. Because these findings apply to every task, they benefit later episodes.

\subsection{Analysis}
\label{sec:results:analysis}

\begin{figure*}[th!]
  \centering
  \includegraphics[width=\linewidth]{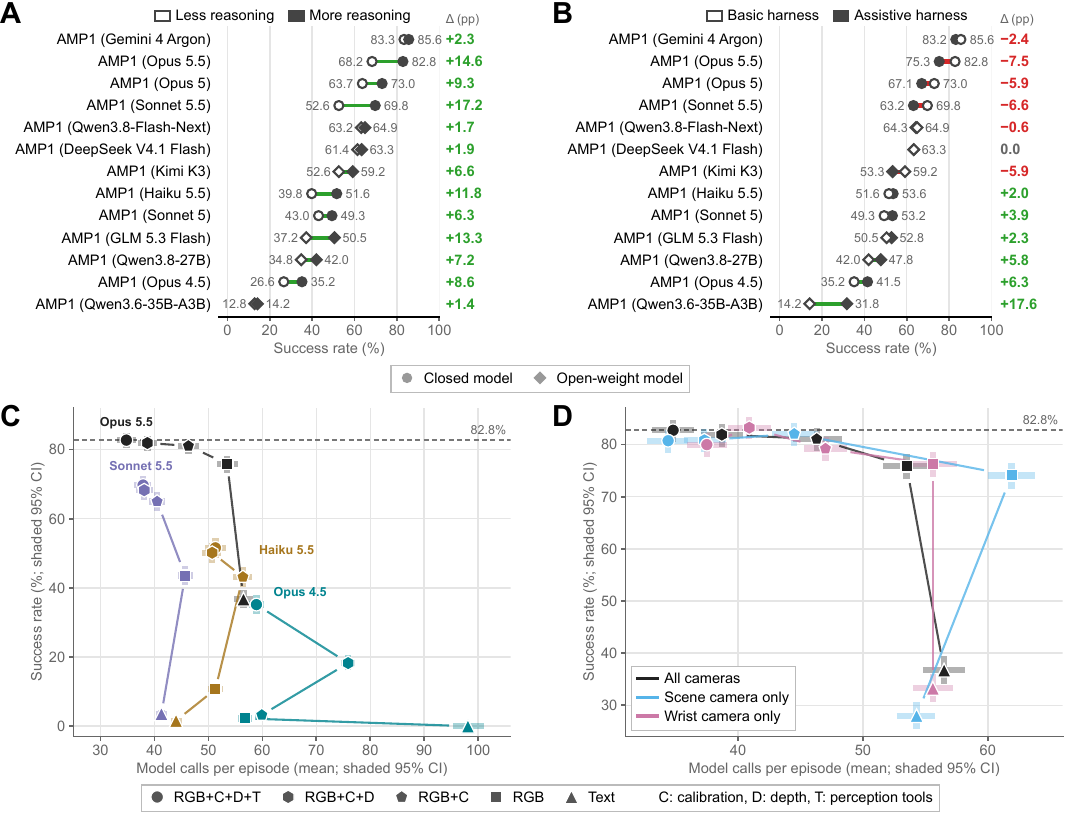}
    \caption{\textbf{Controlled experiments on RoboLab-120.}
        (\textbf{A}) Success rate of 13 models with less and more reasoning, using the basic harness (\cref{sec:results:analysis:model}). 
        (\textbf{B}) Success rate with the basic and assistive harnesses, each model at its more-reasoning setting from A (\cref{sec:results:analysis:motion}).
        (\textbf{C}) Sensory ablation for four Claude models with all cameras, using the basic harness: removing the perception tools (RGB+C+D), then depth (RGB+C), then calibration (RGB), and finally replacing images with textual descriptions (Text).
        (\textbf{D}) Opus~5.5 with different camera constellations.}
  \label{fig:amp_ablation_results}
\end{figure*}

\sysname comprises two interlocking parts: the frontier model and the harness. The latter serves as the former's interface to the world, supplying the sensory inputs and executing the commanded actions.
We separate the contributions of these two parts via controlled experiments on RoboLab-120, varying the sensory input and motion interface on the harness side, and the choice of model and its reasoning effort on the model side (\cref{fig:amp_ablation_results}). We report a number of findings. First, the model matters far more than any change to the harness (\cref{sec:results:analysis:model}). Second, stronger models do better when they command the manipulator directly rather than via a motion interface that adds assistive routines for grasping and placing objects
(\cref{sec:results:analysis:motion}). Third, stronger models cope better with missing sensory input and can compensate for impaired sensor constellations via active perception and additional computation (\cref{sec:results:analysis:perception}). Fourth, 
direct image input is substantially more effective than textual descriptions (\cref{sec:results:analysis:perception}), contrasting with findings reported in prior work~\cite{capx}.

\subsubsection{The choice of model dominates performance}
\label{sec:results:analysis:model}

We hold the harness fixed and vary the model and its reasoning effort (\cref{fig:amp_ablation_results}A). We test 13 models: seven closed models, namely Gemini~4 Argon and six Claude models (Opus~5.5, Opus~5, Opus~4.5, Sonnet~5.5, Sonnet~5, and Haiku~5.5), and six open-weight models, namely three Qwen models (Qwen3.6-35B-A3B, Qwen3.8-27B, and Qwen3.8-Flash-Next), GLM-5.3 Flash, DeepSeek~V4.1 Flash, and Kimi~K3. Each runs at a low and a high reasoning setting, except Qwen3.6-35B-A3B, which we run with reasoning off and at maximum, and Qwen3.8-27B and Qwen3.8-Flash-Next, which we run at low and maximum.

\mypara{The choice of model has the largest effect} Success rates range from 12.8\% with Qwen3.6-35B-A3B to 85.6\% with Gemini~4 Argon. Stronger models consistently outperform weaker or older ones: within the Claude family, the success rate is 82.8\% with Opus~5.5, 51.6\% with the smaller Haiku~5.5, and 35.2\% with the older Opus~4.5. Among open-weight models, Qwen3.8-Flash-Next achieves the highest success rate, 64.9\%.

\mypara{Reasoning helps all models} Claude models gain the most from reasoning, up to 17.2 points for Sonnet~5.5 and 14.6 points for Opus~5.5. Gemini~4 Argon gains only 2.3 points, partly because it has little headroom: it already succeeds in most episodes at low reasoning, and more than half of its remaining failures are in tasks where reasoning is not the limiting factor (Supplementary Materials, \cref{sec:appendix:robolab_failures}).

\subsubsection{Assistive routines help weaker models but hinder stronger ones}
\label{sec:results:analysis:motion}

We compare two motion interfaces for every model on RoboLab-120, each at its more-reasoning setting (\cref{fig:amp_ablation_results}B): the \emph{basic} harness, our default, in which \sysname commands timed end-effector waypoints directly, and an \emph{assistive} harness, which adds routines that plan and execute picks and placements for the agent. Both use the same prompt and perception.

\mypara{Basic harness} \sysname commands motion as a trajectory of timed end-effector poses and gripper commands, and makes all decisions, from where to grasp an object to where to release it. The agent can preview a trajectory before executing it. The harness refuses motions that are unreachable or too fast, but leaves collision checking to the model (\cref{sec:material:motion}).

\mypara{Assistive harness} The assistive harness lets the model delegate grasping and placing through additional routines. The model can request ranked grasp and placement poses. A motion planner executes collision-free picks and placements at the indicated poses. The assistive pick routines check whether the object is held and retry with another grasp if not (\cref{sec:material:motion}). The model still decides which objects to move, in what order, and when the task is complete. It can always take full control via the basic harness, which remains available at all times.

\mypara{Result} The four models with the highest success rates under the basic harness~-- Gemini~4 Argon, Opus~5.5, Opus~5, and Sonnet~5.5~-- all do worse with the assistive harness, by up to 7.5 points (\cref{fig:amp_ablation_results}B). The six models with the lowest success rates all do better, by up to 17.6 points for the weakest model, Qwen3.6-35B-A3B. The three models in between gain nothing.

\mypara{Analysis}
Prior work found that high-level abstractions improved single-turn performance, while multi-turn reasoning could compensate for lower-level interfaces~\cite{capx}. Our results indicate that assistive grasping and placing routines help weaker models but reduce the success rate of stronger models.
The assistive routines are imperfect: a grasp can slip or a placement can miss. Weaker models still benefit from the assistance, because the routines succeed more often than their own direct grasping and placement attempts. For stronger models, the assistive routines are on balance a detrimental distraction, because they hide failures that the models could otherwise immediately observe and correct. For example, an assistive pick routine retries internally before the model can intervene, and reports only whether it succeeded, not what went wrong.
With the basic harness, the model sees the result of every atomic action and can correct mistakes as they happen.

\subsubsection{The effects of sensory impairment}
\label{sec:results:analysis:perception}

RoboLab-120 provides fixed scene cameras and a wrist camera, each with RGB images, depth (D), and calibration (C). On top of these signals, the \sysname harness offers perception tools (T) that build a scene model from the fused point cloud: they segment it into objects, track the objects over the episode, and render them from any viewpoint (\cref{sec:material:perception}). We ablate this input along two axes. The first is how much geometric information each camera provides: starting from the full input (RGB+C+D+T), we remove perception tools (RGB+C+D), then depth (RGB+C), and then calibration (RGB). The second is which cameras \sysname sees through: all of them, the scene cameras only, or the wrist camera only. In a final condition (Text), \sysname receives no images, only textual descriptions of their content, as recommended in prior work~\cite{capx}. The rest of the harness and the prompt are unchanged.

\mypara{Stronger models rebuild missing geometry at the cost of efficiency} With a strong model (Opus~5.5) and all cameras, removing geometric information lowers the success rate only modestly but carries a cost in terms of efficiency (\cref{fig:amp_ablation_results}C). From RGB+C+D+T to RGB+C+D, RGB+C, and RGB, the success rate declines from 82.8\% to 81.9\%, 81.1\%, and 75.8\%, while the mean number of model calls per episode rises from 34.8 to 38.7, 46.3, and 53.5, respectively.
\sysname rebuilds the missing geometry itself. Without the perception tools, it detects objects in the images and lifts them to 3D with depth and calibration. Without depth, it picks matching pixels in zoomed crops of different views and triangulates them through the known calibration. Without calibration, it first estimates the camera parameters, most reliably with the wrist camera, which it can move (a form of active perception~\cite{Bajcsy1988}): it captures images at known gripper poses, matches SIFT features across them~\cite{lowe2004sift}, and fits the focal length, principal point, and camera-to-gripper pose by least squares (Supplementary Materials, \cref{sec:appendix:case_studies:self_calibration}).

\mypara{Weaker models require a more robust sensor constellation} \cref{fig:amp_ablation_results}C compares Opus~5.5 with three weaker models: Sonnet~5.5, Haiku~5.5, and Opus~4.5. While Opus~5.5 is able to compensate for missing sensory input at the cost of efficiency, weaker models are more brittle. The weaker the model, the more sensitive it is to sensory impairment. The success rate of Opus~4.5 drops sharply as soon as the perception tools are taken away. Haiku~5.5 suffers when the perception tools and depth are both withheld. Sonnet~5.5 exhibits material degradation in task completion when the perception tools, depth, and calibration are all removed. The gap between the stronger and weaker models is particularly stark when only RGB input is provided; in this condition, Haiku~5.5 succeeds in only 10.7\% of the episodes, against 75.8\% for Opus~5.5. Weaker models are unable to compensate for sensory impairment even with more reasoning and computation: as sensory input is ablated, \sysname with Opus~5.5 takes substantially more model calls to complete the tasks, while with weaker models the number of model calls does not grow by much as their success rate plummets.

\mypara{Images cannot be replaced by text} Prior work recommended substituting direct camera feeds by textual descriptions of their contents~\cite{capx}.
We test this with four Claude models: Opus~5.5, Sonnet~5.5, Haiku~5.5, and Opus~4.5. In the Text condition, a second instance of the same model describes what the cameras see. \sysname receives these descriptions in place of images. The results are summarized in \cref{fig:amp_ablation_results}C. With Opus~5.5, the success rate drops from 75.8\% with RGB images to 36.8\% with the corresponding textual descriptions. The success rate of weaker models in the Text condition is vanishingly low.
Textual descriptions elide geometric detail, such as whether the gripper is aligned with an object; this has a sharply detrimental effect on task performance.

\mypara{Multiple cameras are not necessary} With a strong model (Opus~5.5), \sysname is robust to removing all but one camera from the sensor constellation (\cref{fig:amp_ablation_results}D). Specifically, in the RGB+C+D+T condition, the agent has an 82.8\% success rate with all cameras, 80.8\% with the scene cameras only, and 80.0\% with the wrist camera only. The spread is confined to three percentage points in all image-based conditions. With only the wrist camera, \sysname compensates by means of active perception~\cite{Bajcsy1988}: it first pulls the arm back to bring more of the scene into view.

\section{Discussion}
\label{sec:discussion}

\subsection{\sysname versus other manipulation policies}
\label{sec:discussion:other_policies}

The policies we compare against fall into four classes: learned policies, skill orchestration, offline program synthesis, and agentic manipulation policies (AMPs). Learned policies, including VLAs and WAMs, map observations to actions end-to-end from robot data. Skill orchestration policies can utilize generalist AI models to sequence pre-existing low-level skills; their performance is thus limited by the reliability of those skills. In offline program synthesis, a generalist AI model writes or refines a control program before deployment; the program can incorporate perception and feedback, but the model does not revise the control strategy as the task unfolds. AMPs keep the generalist model in the control loop, where it can interpret new observations and directly command the manipulator.

\paragraph{Learned policies.} At the time of writing, the strongest learned policies on the RoboLab-120 leaderboard~-- Lakeside, BiMind v0.5, and FLUX~3 Action~-- succeed in 51.0\%, 49.2\%, and 42.9\% of episodes, respectively, in contrast to 85.6\% for \sysname when paired with Gemini 4 Argon and 82.8\% when paired with Opus 5.5. \sysname perceives the environment with the aid of RGB, depth, and camera calibration, whereas typical VLAs and WAMs use RGB alone. The richer sensory input, however, does not explain the gap: when paired with Opus 5.5, \sysname still succeeds in 81.1\% of episodes without depth, and in 75.8\% with RGB input alone (\cref{fig:amp_ablation_results}C). \sysname draws on the broad intelligence of the generalist AI model to decide which objects to move, in what order, and how to react to unanticipated challenges (\cref{sec:results:emergent_behaviors}), whereas a learned policy must acquire its skills from robot data. In addition to their overall poor task performance, specialist manipulation policies are particularly brittle to changes in the phrasing of the task: in the `vague' task specification condition on RoboLab-120, the success rate of FLUX~3 Action drops further to 28.4\%, while \sysname remains at 84.0\% and 79.8\% with Gemini 4 Argon and Opus 5.5, respectively.

\paragraph{Skill orchestration.} Harness VLA~\cite{harness_vla} and HarnessPAI~\cite{harnesspai} delegate motion to a learned VLA policy, so their reliability is bounded by that learned policy.
Like the routines of our assistive harness, but to an even greater degree, a VLA is something of a black box: when it fails, the agent has limited ability to understand the causes and potential mitigations (\cref{sec:results:analysis:motion}). HarnessPAI compensates for this limitation through practice, refining its program over several trial runs of the evaluation tasks. \sysname reaches comparable results zero-shot: 95.2\% against 96.5\% on LIBERO-PRO, and 100.0\% against 96.3\% on robosuite (\cref{sec:results:benchmarks}).

\paragraph{Offline program synthesis.}
On LIBERO-PRO and robosuite, ASPIRE~\cite{aspire} refines a program for each task through repeated execution and debugging, then evaluates the resulting program on held-out configurations. The program can use perception and feedback, but the generalist model does not revise it during deployment. \sysname is evaluated zero-shot on these benchmarks, with the generalist model in the control loop.

\paragraph{Agentic manipulation policies.}
CaP-Agent0~\cite{capx} also keeps a coding model in the control loop. Its harness supplies a learned grasp planner and a library of synthesized geometric utilities, and uses a separate vision-language model to create a textual summary of changes in the scene. The authors found that this textual representation improved performance relative to direct image input. In our experiments, replacing images with textual descriptions is detrimental to task performance. Adding assistive grasping and placing routines also reduces the success rate of stronger models (\cref{sec:results:analysis}).

Astranomy is a concurrent AMP that is based on GPT-6 Astra and was evaluated in simulation on the RoboLab-120 benchmark~\cite{yang2026astranomy}.
Astranomy reaches 74.8\% on RoboLab-120, in comparison to 85.6\% for \sysname with Gemini 4 Argon. Beyond the underlying model, the harnesses differ in three ways. First, Astranomy queries the model for one small end-effector step at a time, whereas \sysname commands a trajectory. Second, Astranomy disables code execution, whereas \sysname acts by means of code. Third, the \sysname harness equips the agent with an editable scene model.
Equalizing the underlying model makes the effect of the harness even more salient: Astranomy reports a 25.3\% success rate on a 30-task subset with Opus~5.5, in comparison to \sysname's 84.7\% with the same model on the same subset of tasks.

\subsection{Outlook}
\label{sec:discussion:deployment}

\sysname has a number of limitations. First, the policy is slow as the robot spends much of each episode waiting for the model to reason. Second, the model cannot reconsider a trajectory while it executes: our evaluations largely concern quasi-static scenes, whereas a moving target or a slipping object may require a response before the model's next turn.

Notwithstanding these limitations, our results indicate that AMPs constitute a promising foundation for robot manipulation. They cannot yet be deployed in mission-critical settings, but neither can the alternatives. Specialist manipulation policies, such as VLAs and WAMs, are comparably fast but unreliable. AMPs are comparably reliable but slow. We see a promising path to practical manipulation in using AMPs as a starting point for developing manipulation systems that are both reliable and fast. A general trend that supports this optimism is that the cost of a given capability level has fallen by roughly an order of magnitude per year~\cite{Gundlach2025,EmbersonRoodman2026}. Thus we can expect the cost of a model call (the `thinking' part of an AMP's execution) to asymptote to zero over time.

Beyond these general trends, the harness itself offers several avenues for acceleration. Our few-shot and continual learning experiments show that the agent can retain useful code and reduce the number of model calls across episodes (\cref{sec:results:practice}). Reusing these routines avoids repeatedly writing code for calibration, geometric computations, and basic motions. Fast local feedback controllers could be synthesized from AMP experience~\cite{Huang2026}. AMP execution traces could also provide supervision for learned policies~\cite{Shen2026:EmbodiedSWE,Liu2026:Guava}. And the model could plan the next motion while the current one executes, thus overlapping thinking with acting.

\section{Materials and Methods}
\label{sec:material}

\begin{widefigure} %
	\centering
	\includegraphics[width=1.0\linewidth]{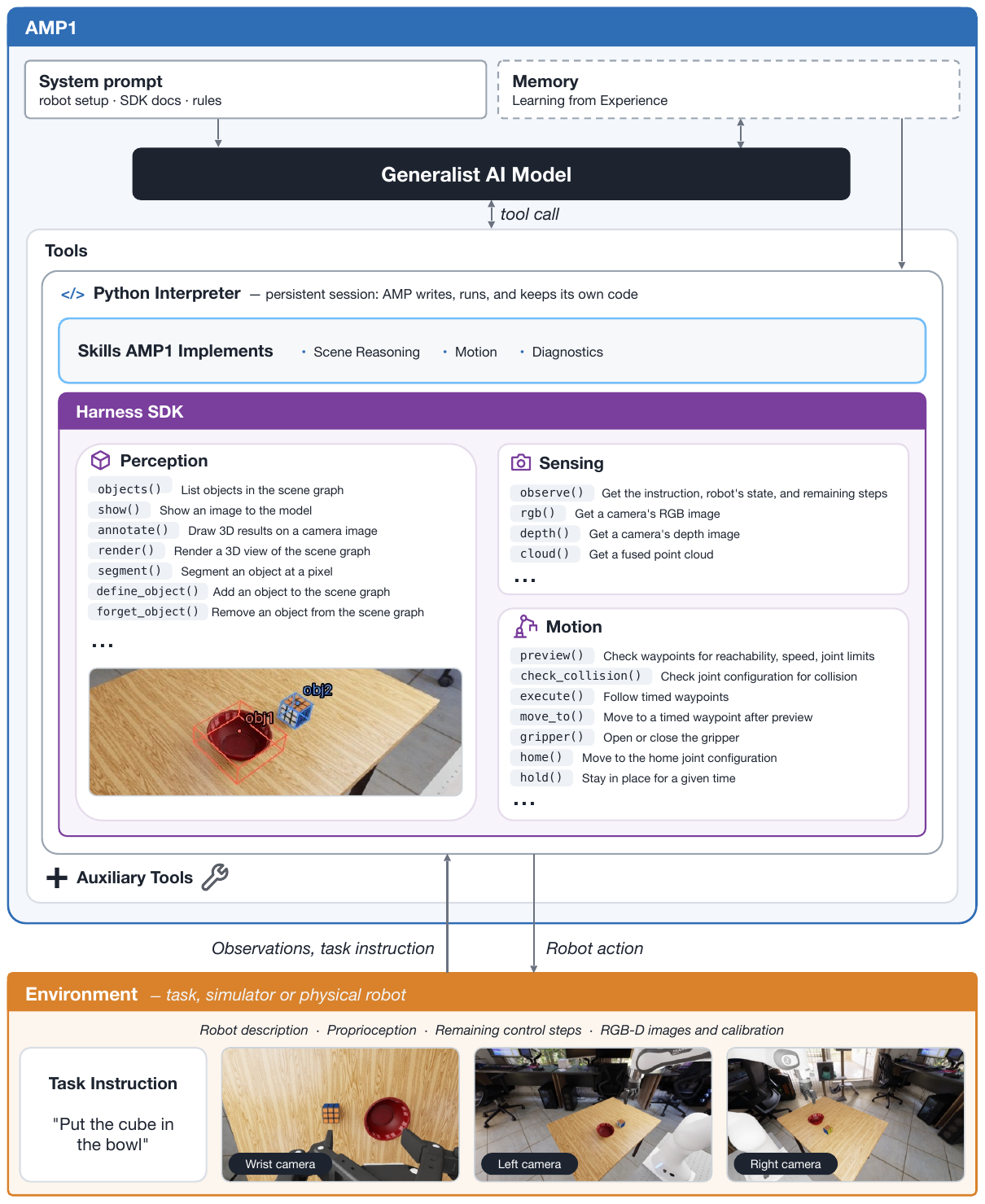}
    \caption{\textbf{\sysname harness.} The generalist AI model controls the robot via Python code that runs in a persistent interpreter. The generalist model builds its own skills on top of the harness SDK. The SDK's functions fall into three groups: sensing, perception, and motion.
    The task instruction and all sensory input reach the model via the harness.
    In addition to the interpreter, the model has a number of auxiliary tools, such as viewing a saved image and terminating the episode (not shown for clarity). The memory (shown dashed) is used when learning from experience (\cref{sec:results:practice}).}
	\label{fig:amp_harness}
\end{widefigure}

\subsection{Harness}
\label{sec:material:amp}
\label{sec:material:harness}

At the core of \sysname is a harness through which a generalist AI model perceives, acts, and reflects (\cref{fig:amp_harness}). The harness comprises (a) an SDK that is called by the model from a persistent Python interpreter (\cref{sec:material:interpreter}); (b) a robot server that serves the SDK's requests: it returns camera images, depth and the robot's state, segments the point cloud into objects and tracks them over the episode, and checks and executes the robot's motion; (c) a route policy that sets which of these requests the harness allows; and (d) reference documentation for the SDK, which is appended to the model's system prompt. Parts (b) and (c) are omitted from \cref{fig:amp_harness} for clarity.

In zero-shot evaluation, everything the model knows about the task and the robot comes through the harness: the instruction, the camera images, depth and calibration, the robot's proprioception, the segmented objects, and the remaining control steps and time. The instruction is read by the model through the SDK in its first turn, together with its first view of the scene.

The harness defines two frames for each end-effector. The end-effector (EE) frame is positioned at the base of the gripper. The tool center point (TCP) frame lies at a fixed offset from the EE frame: its origin, the TCP, is centered between the finger pads, and its axes point along the gripper's approach and closing directions. The model commands motion and defines all grasp geometry at the TCP.

We describe the SDK's functions in three groups: sensing, perception, and motion. Motion comes in two harness variants: the basic harness, our default, in which the model itself commands timed TCP waypoints with gripper commands, and the assistive harness, which adds grasping and placing routines to the basic functions. %

\subsubsection{Sensing}
\label{sec:material:sensing}

\sysname observes the scene and the robot only through SDK calls in the code it implements, and thus the model decides what to observe and when. The only information added to every result the model receives from executing its code are the remaining control steps and wall-clock time. The model may observe the scene arbitrarily often and it may also execute a motion without prior observation and verify its outcome afterwards from the robot's state and camera images.

The model reads the instruction, the maximum number of control steps in an episode, and the steps that remain. Since the episode terminates when its horizon is reached, the model can plan its motion against it. It can also read the robot specification such as its cameras, its workspace, the joint limits and the default configuration, and, for each end effector, its gripper geometry and TCP frame.

The robot's state comprises the measured joint positions $\tilde{q} \in \mathbb{R}^n$ and the poses of the EE and TCP frames in the robot's base frame; we denote the measured TCP pose by $\tilde{T} \in SE(3)$. It also includes the gripper state $\tilde{g} \in [0, 1]$ (0 open, 1 closed) and the last gripper command $g \in \{0, 1\}$.

For each camera, the model can read its RGB image, a depth image subsampled from the RGB image's resolution by a fixed stride, the intrinsics $K$, and the camera's current pose $T_{\mathrm{base,cam}} \in SE(3)$ in the robot's base frame. Helper functions convert between the image and 3D coordinates. One projects a 3D point into the image through $T_{\mathrm{base,cam}}^{-1}$ and $K$, returning its pixel coordinates and its depth along the optical axis, and another inverts this projection for a given pixel and depth. The model can also request a colored point cloud combined from all cameras.

With the rendering functions provided by the harness, the model can compose its own views of the scene from these inputs and the perceived objects. It can draw 3D elements, such as points, boxes, object identifiers, target poses or a planned motion, onto a real camera image by projecting them through $T_{\mathrm{base,cam}}^{-1}$ and $K$, for example to check whether a planned grasp is aligned with the object. It can also render a point cloud, perceived objects, and the robot from any virtual viewpoint, and request a close-up of an object that combines renders from viewpoints around the object. All of these views are built from the robot's own sensor data and the perceived scene model (\cref{sec:material:perception}).

An image enters the model's context only when the model chooses to look at it; that is, when its code displays the image. Every displayed image is also saved. Once the context exceeds a fixed number or total size of images, the oldest are replaced by pointers to their saved files, and the model can reload any of them with a dedicated tool. The model can also keep information from earlier turns, such as images and depth maps, in its persistent interpreter, for example to compare a scene before and after a motion.

\subsubsection{Perception}
\label{sec:material:perception}

The harness maintains a scene model, a set of objects $o_j$, each occupying a region $M_j \subset \mathbb{R}^3$.
It segments a fused point cloud into voxel clusters and extrudes the height map of each cluster into a closed mesh that bounds $M_j$. The union of these regions forms the obstacle region $\mathcal{O} = \bigcup_j M_j$ against which collisions are checked. The generalist AI model can exclude objects (such as the one it is about to grasp) from a collision check. A stereo matching model can be used to estimate depth from stereo images (\cref{sec:material:protocol:real_robot}).

At each update of the scene model, the harness voxelizes the fused point cloud and clusters the voxels into candidate objects as connected components of adjacent voxels. The clustering stops where one object touches another and at sharp color discontinuities. Small fragments, such as an isolated face, are merged into a larger nearby segment of similar color, and close segments of the same color are joined. Segments that remain very small are dropped as noise. Each object is then represented as a closed mesh by extruding its top surface downwards to its lowest observed point. Occluded gaps in the top surface are filled unless the cameras have seen them. Geometry is therefore inferred from depth, while color serves as a secondary criterion that separates touching objects.

The scene model is updated in the background whenever a new camera frame is available and no motion is running, before every collision check, and whenever the generalist AI model requests it. Each object $o_j$ has a unique identifier, a label naming its color and dimensions, its centroid, axis aligned and oriented bounding boxes, and the mesh bounding $M_j$. The generalist AI model can list objects, read their points, size and color, filter them by label, color or size, and sort them by distance to a 3D point.

Each update matches its objects to those in the previous update by a minimal cost assignment, which pairs two objects only if their centroids lie within a distance $d_{\mathrm{m}}$ and their colors and sizes agree within tolerances; a matched object keeps its identifier. When an object is merged with another, split, or keeps its place but no longer matches because its appearance changed, its old identifier resolves to the object that now covers most of its footprint, while an object that moved farther than $d_{\mathrm{m}}$ receives a new identifier. An object that is not seen in an observation update, for example, because the arm occludes it, leaves the scene model, but its identifier is reserved for a fixed time. An object of similar color and size that reappears within $d_{\mathrm{m}}$ of where it was last seen receives the old identifier. Each update also reports what it removed and changed, so that the generalist AI model can follow how the scene model evolves.

The generalist AI model can also correct the scene model. It can grow an object from a single pixel through neighboring points. It can define an object from an axis-aligned or oriented box or from a set of points. And it can replace and remove wrong or spurious objects. Objects that are added by the generalist AI model persist across perception updates and take precedence over the perceived objects they overlap. The system prompt warns the generalist AI model that perception can merge touching objects, split one object in two, or fill in hidden parts incorrectly. It instructs the generalist AI model to check the perceived objects against the camera images and, when they are incorrect, to repair the scene model with these tools rather than compensate for the error in its plans. After every motion, the generalist AI model is instructed to update the scene model and look at the cameras.

\subsubsection{Motion}
\label{sec:material:motion}

The generalist AI model writes code that produces a trajectory of TCP poses with gripper commands, given as timed waypoints ${w_i = (T_i, t_i, g_i)}$, $i = 1, \dots, N$. Each waypoint consists of (a) a desired TCP pose $T_i = (p_i, R_i) \in SE(3)$ in the robot's base frame, with position $p_i \in \mathbb{R}^3$ and orientation $R_i \in SO(3)$, (b) the desired arrival time $t_i$,  and (c) a gripper command $g_i \in \{0, 1\}$ (open or closed). The model typically constructs each waypoint's TCP pose relative to a reference it has measured, such as the current TCP pose, an object's centroid, or the pose of another waypoint, using helper functions that either build a pose from a position and an approach direction (for example, top-down at a given yaw) or shift a pose along its own axes. The harness provides no function that chooses grasps or placements; those decisions are left to the generalist model. For a bimanual robot, the model generates a sequence of waypoints for each arm.

From the waypoints, the harness computes a joint space trajectory $\tau\colon [0, t_N] \to \mathcal{C}$, where $\mathcal{C} \subset \mathbb{R}^n$ is the configuration space of the robot's $n$ joints, with configurations $q \in \mathcal{C}$. Starting from the current TCP pose $T_0 = (p_0, R_0) = f(q_0)$ at $t_0 = 0$, where $f\colon \mathcal{C} \to SE(3)$ is the forward kinematics, the harness interpolates between consecutive waypoints, treating position and orientation separately rather than as a single screw motion (the decoupled path of Sec.~9.2.1 in \cite{LynchPark2017}). More precisely, for $t$ in the segment $[t_{i-1}, t_i]$ between waypoints $i-1$ and $i$, the interpolated TCP pose is $T(t) = (p(t), R(t))$ with ${p(t) = p_{i-1} + \sigma(u)\,(p_i - p_{i-1})}$ and ${R(t) = R_{i-1} \exp\!\big(\sigma(u) \log(R_{i-1}^{\top} R_i)\big)}$. Here, $p_i$ and $R_i$ are the position and orientation of the waypoint $i$, ${u = (t - t_{i-1}) / (t_i - t_{i-1}) \in [0, 1]}$ is the normalized time within the segment, and $\sigma$ is a quintic polynomial time scaling that minimizes jerk.

At every control step $k$ of duration $\Delta t$, the harness computes ${q_k = \tau(k\,\Delta t)}$ by damped least squares inverse kinematics of ${T(k\,\Delta t) = (p(k\,\Delta t), R(k\,\Delta t))}$, seeded with $q_{k-1}$. The numerical inverse kinematics solver always returns the best configuration $q_k$ it can find for an interpolated pose $T(k\,\Delta t)$, but when the pose is unreachable or the solver does not converge, $q_k$ reaches it only approximately. The harness therefore validates the trajectory and executes it only if the following conditions hold at every step $k$. First, the position and orientation errors between $f(q_k)$ and $T(k\,\Delta t)$ must lie within specified tolerances. Second, each joint's speed must stay below its speed limit. Approaching near the end of a joint's range of motion or near a singularity only sends a warning to the model. Additionally, the model can also test whether $\tau$ lies in $\mathcal{C}_{\mathrm{free}} = \mathcal{C} \setminus \mathcal{C}_{\mathrm{obs}}$, where $\mathcal{C}_{\mathrm{obs}} = \{q \in \mathcal{C} \mid \mathcal{A}(q) \cap \mathcal{O} \neq \emptyset\}$, by checking configurations it samples from $\tau$. Here, $\mathcal{A}(q) \subset \mathbb{R}^3$ is the region occupied by the robot at configuration $q$, which the harness approximates by spheres covering the arm and the gripper, and $\mathcal{O}$ is the obstacle region formed by the perceived objects (\cref{sec:material:perception}). A refused trajectory is returned with the reason and the affected steps, so that the model can repair it. Before planning a motion, the model can also test whether a single pose is reachable, with the same tolerances.

Once the entire trajectory has passed these checks, the harness executes it by sending the robot the computed absolute joint position targets, one $q_k$ per control step, each with the current gripper command.

\subsection{Python Interpreter}
\label{sec:material:interpreter}

The agent acts primarily through a Python interpreter, which executes the code implemented by the model in a kernel that persists during the entire episode. The kernel keeps imports, variables, and the functions the model defines between calls, so that the information it generates in one turn can be reused in later ones. Each call returns the code's printed output, any error with its traceback, and the images the code displays, followed by the remaining control steps and wall-clock time. The model reads the result at the start of its next turn. The model has a number of auxiliary tools in addition to the interpreter, such as reloading an image saved earlier, and ending the episode.

In the basic harness, the motion interface only previews, checks, and executes waypoints and gripper commands. Everything in between~-- from choosing a grasp to measuring a container, pushing an object, or recovering from a refused trajectory~-- the model implements itself, combining SDK calls with its own geometry code.
In zero-shot evaluation, each episode starts from a new sandboxed kernel, so the model writes its skills from scratch every time.
\Cref{sec:appendix:worked-example} in the Supplementary Materials shows how the model builds such skills during one episode, turn by turn.
When learning from experience, the model can decide to save its skills into the memory store $\mathcal{M}$ (\cref{sec:results:practice}). 

\subsection{RoboLab}
\label{sec:material:protocol}
\label{sec:material:protocol:robolab}

RoboLab-120 comprises 120 language-specified pick-and-place tasks simulated in Isaac Sim~\cite{robolab}. Each task is posed with three variants of its instruction: default, vague, and specific.

\mypara{Embodiment setup}
The robot is a simulated Franka Panda arm in the DROID configuration, with a Robotiq 2F-85 gripper, mounted at a table. Physics runs at 120\,Hz and control at 15\,Hz. Each action specifies seven absolute joint positions and a gripper command. Two fixed over-the-shoulder cameras and a wrist camera render $1280 \times 720$ images, with horizontal fields of view of about $104^\circ$ and $88^\circ$, respectively. 

\mypara{Evaluation protocol}
Following the benchmark's published protocol, we run 10 episodes per task (i.e., 1,200 in total) for each of the three instruction variants. All episodes of a task start from the same initial scene, so they differ only through the policy and simulator nondeterminism. An episode succeeds at the first step at which the task's success condition holds, and fails if the task's time limit, between 20 and 300\,s of simulated time, expires first. The simulation is paused while the agent reasons, perceives, and plans, so only executed motion counts against this limit. The agent receives the robot model (kinematics, joint limits, and collision geometry). Depth is rendered without noise, and camera intrinsics and extrinsics are exact. 

\subsection{CaP-Gym}
\label{sec:material:protocol:capgym}

CaP-Gym~\cite{capx} wraps three simulation benchmarks behind a common interface: LIBERO-PRO~\cite{libero_pro}, robosuite~\cite{robosuite}, and BEHAVIOR-1K~\cite{li2024behavior1k}. Following the published protocol, we evaluate three of the four LIBERO-PRO suites (LIBERO-Object, LIBERO-Goal, and LIBERO-Spatial), all seven robosuite tasks, and two BEHAVIOR-1K tasks.

\mypara{Embodiment setup}
The embodiment differs across benchmarks. LIBERO-PRO and robosuite simulate a Franka Panda arm with its standard parallel-jaw gripper in MuJoCo~\cite{mujoco}. Physics runs at 500\,Hz and control at 20\,Hz. Each action specifies seven absolute joint positions and a gripper command. In robosuite, the two bimanual tasks use two opposing Panda arms. (The `spill wipe' task replaces the gripper with a wiping tool.) BEHAVIOR-1K simulates a Galaxea R1Pro mobile manipulator in OmniGibson on Isaac Sim, with a holonomic base, a four-joint torso, and two seven-joint arms with parallel-jaw grippers, controlled at 30\,Hz. LIBERO-PRO uses a fixed camera and a wrist camera at $800 \times 512$. robosuite uses a single camera at $512 \times 512$. BEHAVIOR-1K uses a head camera and two wrist cameras at $512 \times 512$.

\mypara{Evaluation protocol}
We follow CaP-Gym's environment protocol, including its tasks, controllers, cameras, step horizon, and success checks. LIBERO-PRO perturbs each of its 30 tasks in two ways, swapping object positions or changing the instruction to name a different object, yielding 60 tasks with 50 trials each (i.e., 3,000 episodes). Robosuite runs 100 trials per task (i.e., 700 episodes), with object positions randomized at every reset. BEHAVIOR-1K runs 25 trials per task (i.e., 50 episodes), each starting from a fixed, pre-sampled scene instance. Each episode is limited by the environment's step horizon: 8,000 steps for LIBERO-PRO and 1,000--5,000 for robosuite. BEHAVIOR-1K has no step horizon and is bounded only by wall-clock time. As in RoboLab, the simulation is paused while the agent reasons, so only executed motion counts against the horizon.
Depth and camera calibration are part of CaP-Gym's observations. Depth is rendered without noise, and camera intrinsics and extrinsics are exact.

\subsection{RoboDojo}
\label{sec:material:protocol:robodojo}

RoboDojo~\cite{robodojo} comprises 42 bimanual manipulation tasks simulated in Isaac Sim, grouped into five categories: generalization (12 tasks), memory (6), long-horizon (8), precision (8), and open-ended (8).

\mypara{Embodiment setup}
The robot consists of two ARX X5 arms with six joints and a parallel-jaw gripper on each, mounted at fixed bases on a table. Physics runs at 250\,Hz and control at 25\,Hz. Each action specifies six absolute joint positions and a gripper command per arm. A head camera and two wrist cameras render $640 \times 480$ images, with horizontal fields of view of about $96^\circ$ and $78^\circ$, respectively.

\mypara{Evaluation protocol}
We conduct our few-shot learning and continual learning experiments on RoboDojo (\cref{sec:results:practice}).
The benchmark defines three public evaluation seeds (0, 1, and 2), each specifying 50 predetermined layouts per task with variations in objects, their positions, backgrounds, and clutter. We use seed 0 for experience collection. In few-shot learning, practice is run five times per task, and the run with the highest success rate in seed 1 is selected for each task. In both few-shot and continual learning, we report success rates on seed 2.
We calculate the success rate as the public leaderboard does. Each category's success rate is the mean over its tasks, and the overall success rate is the mean over the five categories. Confidence intervals are 95\% percentile bootstrap intervals over tasks, resampled within each category.
As in RoboLab, simulation is paused while the agent reasons.
We follow the input protocol of the GPT-6 Astra entry on the public leaderboard, submitted by the RoboDojo authors~\cite{gpt6_robodojo}. For camera data, the agent receives only RGB images, without depth or camera calibration. Each task instruction is accompanied by a task recipe that describes the scene and gives a scoring rubric. Four tasks also have task notes, which we summarize here. In `imitate sorting sequence', the robot must hold still through the demonstration and then reproduce the observed order. In `make kong', it must wait for the opponent's discard and then declare the kong in two motions. In `match and pick from conveyor', the belt does not stop, so the grasp must be aimed where the object will be. In `play tic-tac-toe', the robot must hold still during the opponent's turns, play to win or block, and place all five pieces.

\subsection{Real-world trials}
\label{sec:material:protocol:real_robot}

\mypara{Harness}
We made minimal adjustments to the harness SDK's reference documentation, replacing simulation references with their real-world equivalents.
To support recovery from errors (RfE), the harness includes two SDK functions: one queries the arm's state, the other requests robot error recovery. When a Franka collision reflex interrupts motion, the harness reports the error to the model. The model can generate code that calls these functions. The harness executes the code and returns the results. Following successful recovery, the model can command a revised trajectory.

\mypara{Embodiment setup}
We use the DROID setup \cite{droid}, comprising a 7-DoF Franka Research~3 arm with a Robotiq 2F-85 gripper, a wrist-mounted ZED Mini attached using the official DROID fixture, and two ZED~2i cameras providing left and right shoulder views. The shoulder cameras are positioned to approximate the baseline policies' training-time camera configuration, with each baseline using its expected views. 
All cameras operate at 30\,fps with an output resolution of $1280 \times 720$.
When depth is needed, we use a pretrained S\textsuperscript{2}M\textsuperscript{2} stereo model to obtain metric depth from the ZED cameras~\cite{s2m2}.
Each method uses its expected views: $\pi_{0.5}$ takes the left shoulder and wrist cameras as visual observations~\cite{pi05}, whereas FLUX 3~\cite{flux3a} and \sysname additionally use the right shoulder camera. 

\mypara{Evaluation protocol}
We evaluate each method on 12 complex real-world tasks. To assess robustness to visual and hardware variation, we use three instances of the platform that differ in scene background and hardware characteristics, including camera calibration, cabling, and component wear. For each task and method, we conduct 9 trials, with 3 trials on each platform instance.
The instructions are issued in free-form natural language and span single-skill pick-and-place as well as multi-stage instructions that require subgoal sequencing. 
For fair comparison, we build a scene restoration tool that ensures that, for every trial of each task, the initial object poses (projected into the workspace relative to the robot base) and the set of objects are consistent across all methods.
The initial scene and the language instruction for each task are detailed in Supplementary Materials.
All methods are evaluated under the same 5-minute execution time budget (excluding reasoning time).
If the time limit is exceeded or an uninstructed object is erroneously grasped and placed, the trial is considered unsuccessful. For the baseline deployments and \sysname without RfE, a robot controller error that interrupts motion also terminates the trial. For \sysname with RfE, a recoverable controller error does not by itself terminate the trial; the agent can request error recovery through the harness and, if recovery succeeds, continue the same trial within the remaining execution-time budget.

\subsection{Compute}
\label{sec:material:protocol:compute}
Closed-weight models are queried through their providers' APIs. Open-weight models are self-hosted with SGLang. Qwen3.6-35B-A3B and Qwen3.8-27B run on H100 GPUs, the others on B200 GPUs. We do not set sampling parameters such as temperature and top-$p$, so each model uses its provider's defaults. In simulation, multiple episodes run in parallel on a node with eight H100 GPUs, which hosts the simulator together with the harness and the agent's code execution. On the real robot, the harness runs on a local workstation with two RTX Pro 6000 GPUs. The baseline policies are also evaluated on the same platform.

\references

\clearpage
\section{Supplementary Materials}
\renewcommand{\thesubsection}{S\arabic{subsection}}
\renewcommand{\thesubsubsection}{S\arabic{subsection}.\arabic{subsubsection}}
\renewcommand{\thefigure}{S\arabic{figure}}
\renewcommand{\thetable}{S\arabic{table}}
\renewcommand{\theequation}{S\arabic{equation}}
\setcounter{subsection}{0}
\setcounter{figure}{0}
\setcounter{table}{0}
\setcounter{equation}{0}

\subsection{Real-world trials}
\label{sec:appendix:real_world_benchmark}

This section complements the embodiment setup and the evaluation protocol described in~\cref{sec:material:protocol:real_robot} with more details of our real-world experiments.
\cref{fig:platforms} shows the three instances of our DROID-based platform. 
They use the same arm, gripper, and camera models, and differ in tabletop background and in unit-level hardware (camera calibration, cabling, component wear).
Every method is run for 3 trials per task on each instance, i.e., 9 trials per task and method.
\cref{sec:appendix:fig:tasks} shows the initial scene and the language instruction of each of the 12 tasks. 

\begin{widefigure}
  \centering
  \includegraphics[width=\textwidth]{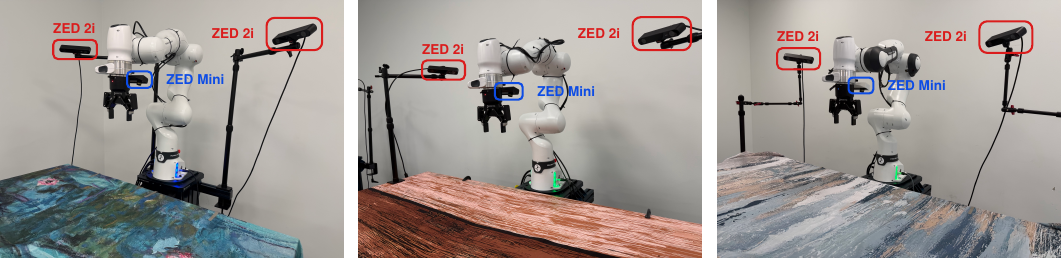}
  \caption{\textbf{Three instances of our DROID-like physical evaluation platform.}
  The instances use different tabletop backgrounds, among other unit-level differences. The two shoulder cameras
(ZED~2i) are annotated in red and the wrist camera (ZED~Mini) in blue.}
  \label{fig:platforms}
\end{widefigure}

In the real-world evaluation results (\cref{tab:results:realworld}), 
the time to success of \sysname is split into \emph{execution} (Exec.)\ and \emph{reasoning} (Reason.).
\emph{Execution} time covers only the intervals in which the robot is carrying out the agent's commanded actions; 
\emph{reasoning} time accumulates the agent's inference, the execution of its own code, the camera capture, and the 3D perception processing performed after each motion.
In addition, we annotate each of the 12 tasks with three kinds of features, shown in the columns \emph{Manip.}, \emph{Horizon}, and \emph{Reasoning}. 
These features let us relate a method's success to the demands of a task; we explain each of them in more detail below:

\begin{itemize}
    \item \textbf{Manipulation} describes the kinds of actions a task requires. 
    A task is \emph{prehensile} if it can be solved by grasping and placing objects alone. 
    A task is labeled \emph{non-prehensile} if completing it requires non-prehensile actions, such as pushing, sliding, or nudging an object, so that purely prehensile manipulation cannot solve it.
    A task can have both labels when completing it requires both kinds of motion. 
    For example, in the task ``Put the bowl on the plate'', the bowl is flipped and no initial grasp is feasible, so the robot first pushes it to the edge of the bin and then grasps it.
    
    \item \textbf{Horizon} separates \emph{single-step} and \emph{multi-step} tasks. 
    A single-step task usually involves manipulating only one target object (e.g., picking and placing one object). 
    A multi-step task requires completing multiple subtasks in sequence, 
    usually by manipulating multiple objects one after another (e.g., putting all the red things in the bin).

    \item \textbf{Reasoning} lists the kinds of reasoning a task requires to decide what to manipulate and how. Multiple kinds may be called for per task:
    
    \begin{itemize}
      \item \emph{Spatial reasoning}: understanding a spatial relation, e.g., ``to the left of''.
      
      \item \emph{Semantic grounding}: identifying objects from their color or category, e.g., ``red things'' or ``canned foods''.
      
      \item \emph{Exclusion}: acting on all objects of a class except a named one, e.g., ``all the tools except the hammer''.
      
      \item \emph{Ordering}: deciding the order in which to act, e.g., comparing cup sizes to nest them.
      
      \item \emph{Affordance}: an enabling action is needed to make the next action feasible, 
      e.g., pushing a flipped bowl to the bin edge before grasping it.
  
      \item \emph{Memory}: using information that is no longer visible, e.g., remembering which cup hides the yellow cube in order to uncover it.
    \end{itemize}
      
  \end{itemize}

\begin{widefigure}
  \centering
  \includegraphics[width=\textwidth]{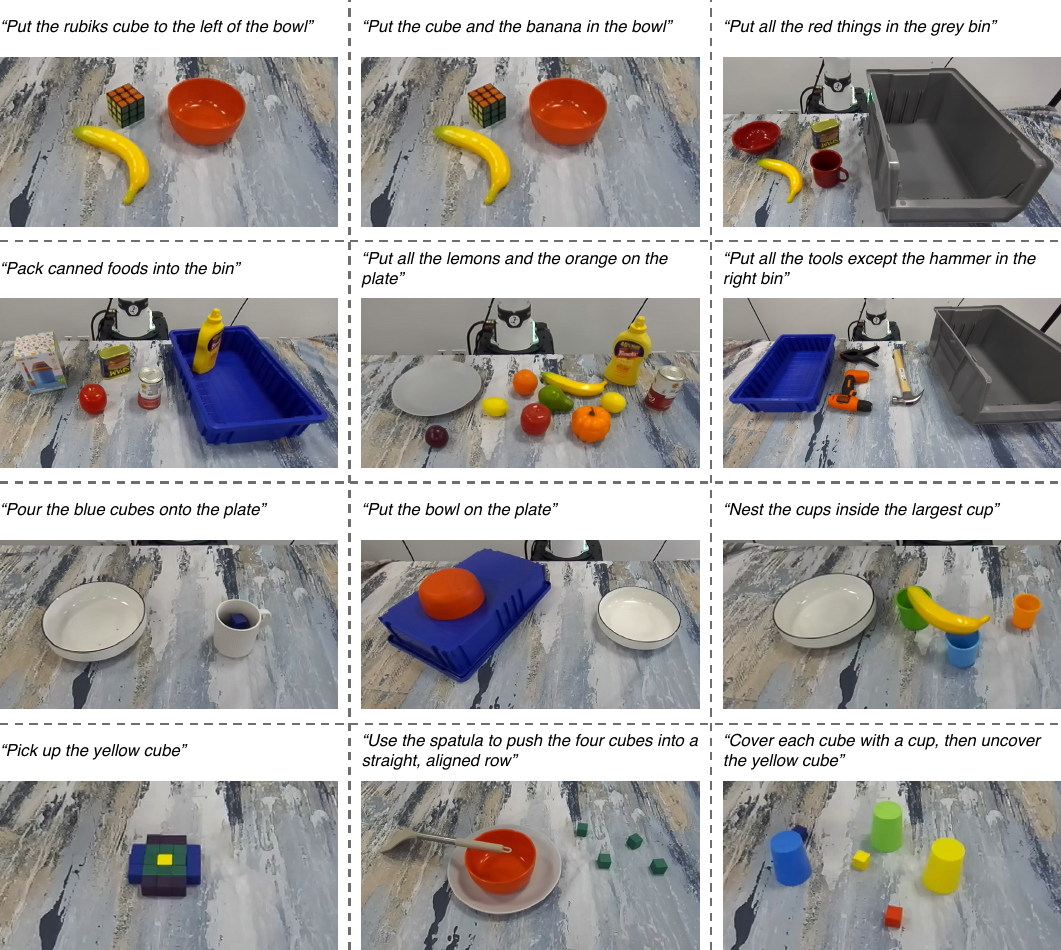}
  \caption{\textbf{The 12 real-world evaluation tasks.}
  Each panel shows the initial scene of one task, labeled with its language instruction. 
  Note that the images only show initial conditions from one platform instance, whereas the evaluations are conducted on all three platform instances shown on \cref{fig:platforms} independently.}
  \label{sec:appendix:fig:tasks}
\end{widefigure}

Complementing the baseline comparison in \cref{tab:results:realworld}, we ablate access to the two error-recovery SDK functions in \cref{tab:results:rfe}.
Without these functions, a robot controller error that interrupts motion terminates the trial under our evaluation protocol. With recovery from error (RfE), the agent can request recovery through the harness and continue the same trial if recovery succeeds. \sysname succeeds in 104 of 108 trials (96.3\%) with RfE and 94 of 108 trials (87.0\%) without RfE. The reported average time to success is similar across conditions, although each average includes only the successful trials in that condition.
The gain is largest on tasks involving close contact between the gripper or held object and the scene, where collisions are more likely (e.g., ``Pack canned foods into the bin'', from 3\,/\,9 to 9\,/\,9).

\begin{table*}[!tp]
    \centering
    \small
    \caption{\textbf{Ablation of recovery from error (RfE) in the real-world evaluations.}
    Success (successes/trials) and average time to success of \sysname and of a variant without RfE (w/o RfE) on the 12 tasks of \cref{tab:results:realworld}.
    In both conditions, the Franka arm's collision reflex interrupts motion. Without RfE, we terminate the trial; with RfE, the agent can autonomously request error recovery through the harness and continue if recovery succeeds. Time to success is averaged over successful trials only.
    }
    \label{tab:results:rfe}
    \setlength{\tabcolsep}{2pt}
    \setlength{\aboverulesep}{0pt}
    \setlength{\belowrulesep}{0pt}
    \renewcommand{\arraystretch}{1.3}
    \begin{tabular*}{\textwidth}{@{\extracolsep{\fill}}>{\raggedright\arraybackslash}p{\dimexpr\textwidth-262pt\relax}||cc|ccc|ccc@{}}
        \toprule
        \multirow{3}{*}{Task Instruction} & \multicolumn{2}{c|}{Success} & \multicolumn{6}{c}{Average Time to Success (seconds)} \\
        \cmidrule(lr){2-3}\cmidrule(lr){4-9}
        & \sysname{} & \sysname{} & \multicolumn{3}{c|}{\sysname{}} & \multicolumn{3}{c}{\sysname{} (w/o RfE)} \\
        \cmidrule(lr){4-6}\cmidrule(lr){7-9}
        & & (w/o RfE) & Exec. & Reason. & Total & Exec. & Reason. & Total \\
        \midrule
        Put the rubiks cube to the left of the bowl & 9 / 9 & 9 / 9 & 22 & 72 & 94 & 21 & 67 & 88 \\
        Put the cube and the banana in the bowl & 9 / 9 & 9 / 9 & 44 & 138 & 182 & 50 & 151 & 201 \\
        Put all the red things in the grey bin & 9 / 9 & 8 / 9 & 46 & 146 & 192 & 54 & 156 & 210 \\
        Pack canned foods into the bin & 9 / 9 & 3 / 9 & 54 & 154 & 208 & 43 & 114 & 157 \\
        Put all the lemons and the orange on the plate & 9 / 9 & 8 / 9 & 75 & 115 & 190 & 65 & 116 & 181 \\
        Put all the tools except the hammer in the right bin & 9 / 9 & 8 / 9 & 102 & 301 & 403 & 61 & 232 & 293 \\
        Pour the blue cubes onto the plate & 7 / 9 & 7 / 9 & 39 & 196 & 235 & 63 & 229 & 292 \\
        Put the bowl on the plate & 7 / 9 & 8 / 9 & 74 & 410 & 484 & 93 & 459 & 552 \\
        Nest the cups inside the largest cup & 9 / 9 & 9 / 9 & 74 & 188 & 262 & 74 & 218 & 292 \\
        Pick up the yellow cube & 9 / 9 & 9 / 9 & 76 & 244 & 320 & 67 & 294 & 361 \\
        Use the spatula to push the four cubes into a straight, aligned row & 9 / 9 & 7 / 9 & 126 & 475 & 601 & 70 & 392 & 462 \\
        Cover each cube with a cup, then uncover the yellow cube & 9 / 9 & 9 / 9 & 90 & 136 & 226 & 93 & 186 & 279 \\
        \midrule
        Total & 96.3\% & 87.0\% & 69 & 211 & 280 & 64 & 219 & 283 \\
        \bottomrule
    \end{tabular*}
\end{table*}

\subsection{Further illustration of emergent behaviors}
\label{sec:appendix:emergent_behaviors}

\cref{fig:emergent_behavior_appendix_1,fig:emergent_behavior_appendix_2} show the remaining behaviors described in \cref{sec:results:emergent_behaviors} that did not fit in \cref{fig:emergent_behavior}.

\begin{widefigure} %
	\centering
    	\includegraphics[width=\linewidth, trim=0pt 0pt 0pt 0pt, clip]{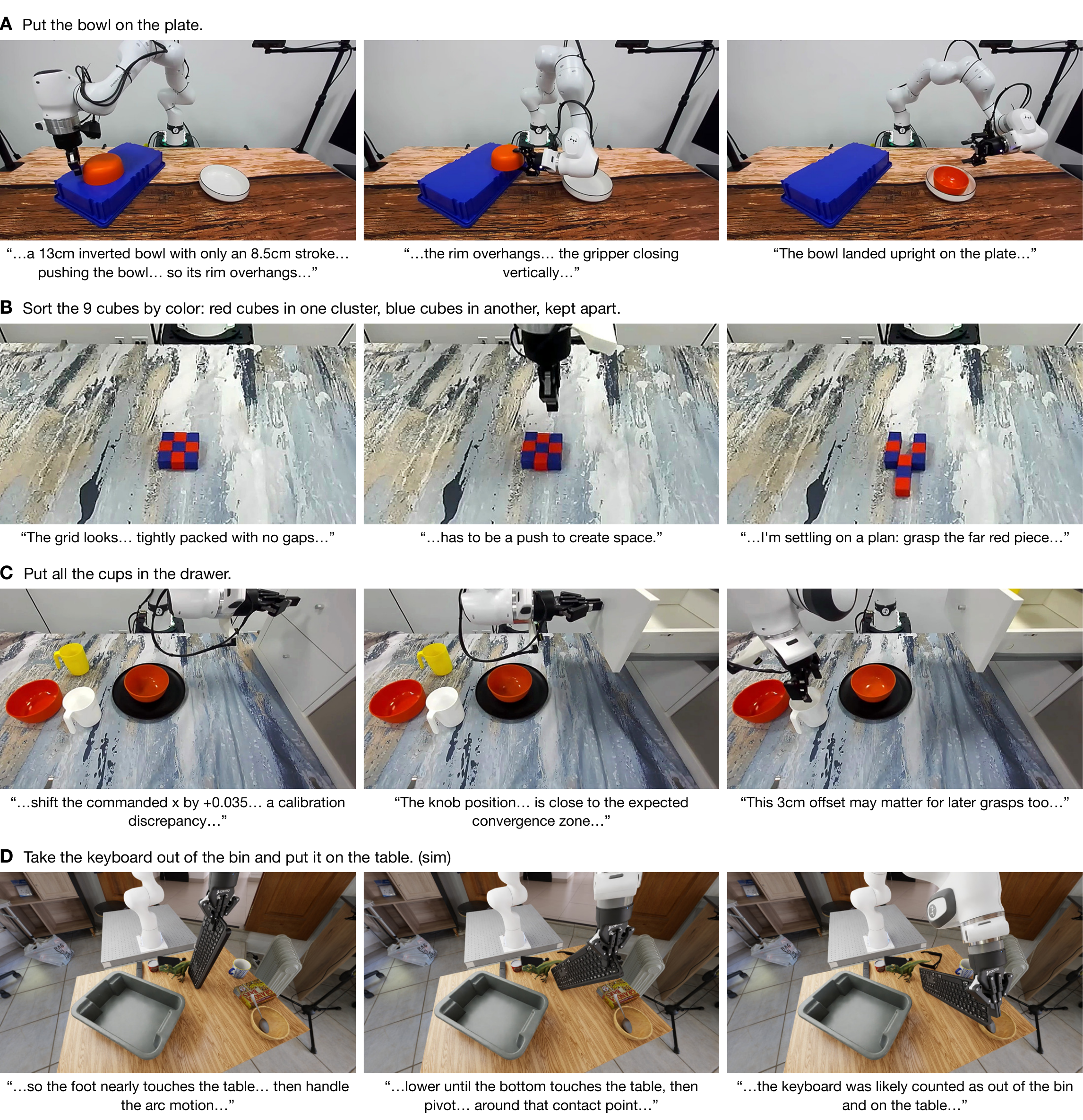}

	\caption{\textbf{Additional emergent behaviors in \sysname (continued from \cref{fig:emergent_behavior}).} Four more episodes in which \sysname works around an unforeseen obstacle with a strategy its prompt does not describe: (\textbf{A} to \textbf{C}) on the real robot, (\textbf{D}) in RoboLab simulation. Quotes under each image are from \sysname's thinking trace. (\textbf{A}) The bowl is too wide to grip and sits flat on a bin, so \sysname slides it until its edge sticks out past the bin, then grips that edge from the side (cf. \cref{fig:emergent_behavior}B). (\textbf{B}) The cubes start in a tightly packed grid that leaves no room for the fingers, so \sysname first pushes a column of the grid to make space. (\textbf{C}) The gripper does not line up with the knob in the wrist camera, which suggests a calibration error, so \sysname closes its gripper to measure the offset and corrects for it in the later subtasks. (\textbf{D}) The keyboard is too long to set down flat in one motion, so \sysname first rests one end on the table, then lowers the other along an arc.}
	\label{fig:emergent_behavior_appendix_1}
\end{widefigure}

\begin{widefigure} %
	\centering
    	\includegraphics[width=\linewidth, trim=0pt 0pt 0pt 0pt, clip]{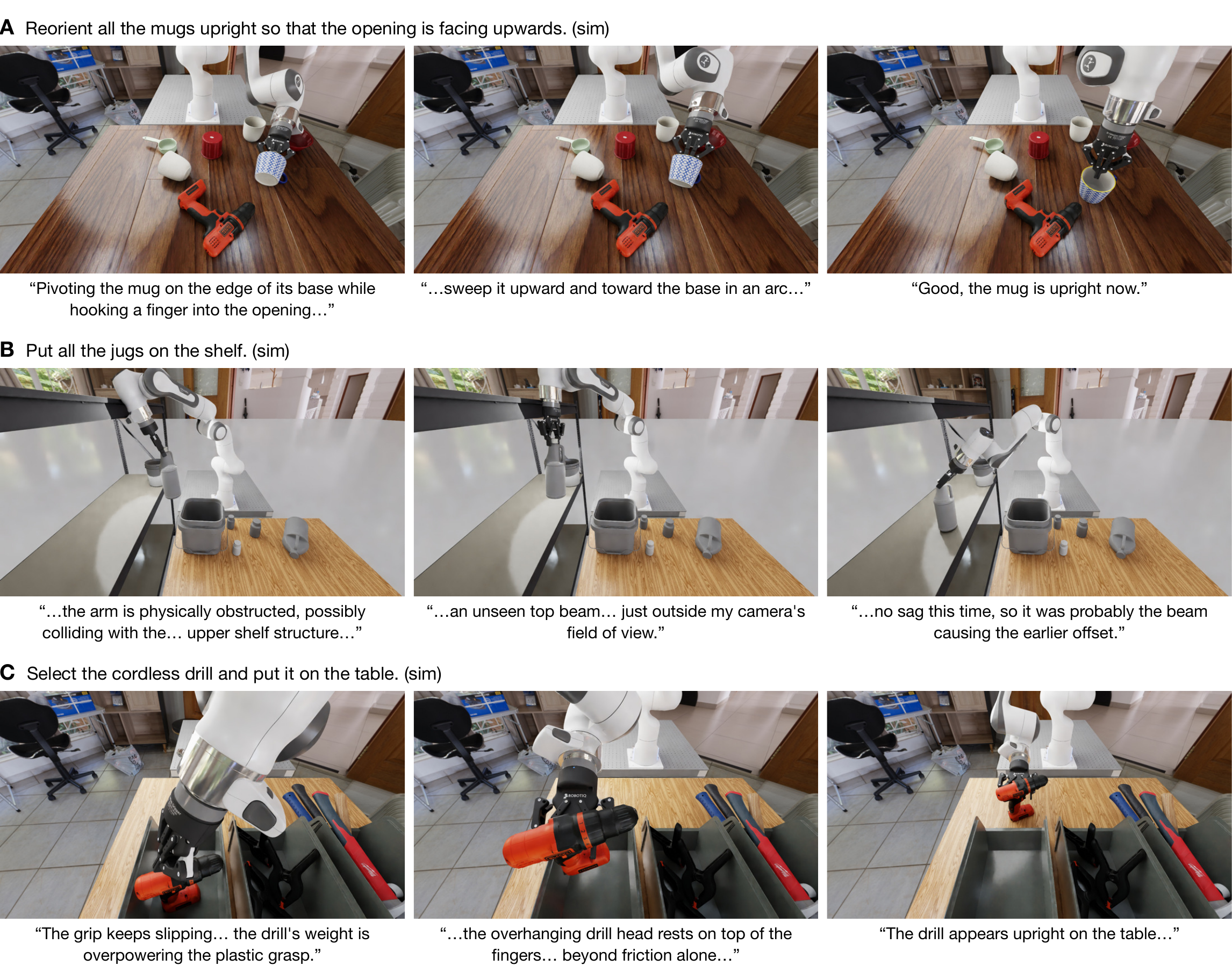}

	\caption{\textbf{Additional emergent behaviors in \sysname (continued).} Three more episodes in which \sysname works around an unforeseen obstacle with a strategy its prompt does not describe, all in RoboLab simulation. Quotes under each image are from \sysname's thinking trace. (\textbf{A}) The mug lies on its side, so \sysname hooks its closed fingers into the mug's opening and pivots it upright. (\textbf{B}) While reaching toward the shelf, the arm is blocked by a structure none of its cameras can see, so \sysname infers an upper shelf beam from the blocked motion and keeps its arm below it. (\textbf{C}) The drill keeps slipping from friction grasps, so \sysname switches to gripping the handle so that the drill head rests on top of the fingers.}
	\label{fig:emergent_behavior_appendix_2}
\end{widefigure}

\subsection{How \sysname reasons and acts: a turn by turn example}
\label{sec:appendix:worked-example}

\begin{figure*}[p]
      \centering
      \includegraphics[width=\textwidth,height=0.9\textheight,keepaspectratio]{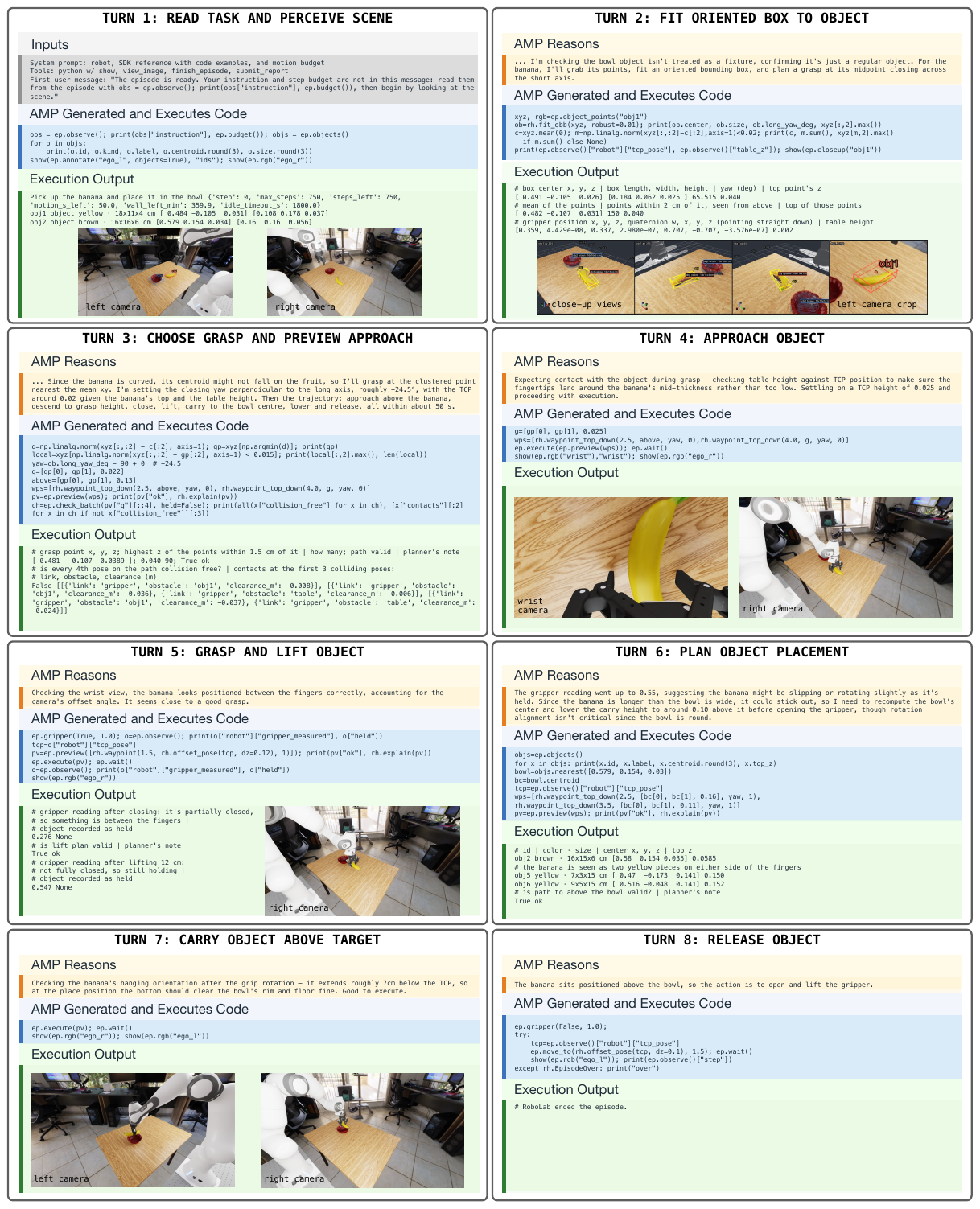}
      \caption{\textbf{One episode, turn by turn.}
          \sysname with Claude Opus~5.5 at high reasoning effort and the basic harness solves the \texttt{BananaInBowlTask} RoboLab task (``Pick up the banana and place it in the bowl''). Each panel is one turn: \sysname reasons (yellow), implements and runs Python code (blue), and receives the execution output (green), which it reads at the start of the next turn.}
      \label{fig:one_episode}
\end{figure*}

\Cref{fig:one_episode} follows one complete episode, on the RoboLab \texttt{BananaInBowlTask}, to demonstrate how \sysname reasons and acts. The agent runs Claude Opus~5.5 at high reasoning effort with the basic harness. It commands the manipulator by means of timed waypoints with gripper commands. It solves the task in eight turns, in each of which it writes and runs Python code.

The episode exhibits a generic decision loop that repeats across tasks. \sysname gathers the information needed for the decision, builds its own geometric model of the object to manipulate from the perceived points, previews its motion before executing it, and checks the outcome in images and sensor readings before the next decision.

The first turn starts from a) the system prompt, which describes the robot, the SDK and the rules of the episode, b) the tool definitions, and c) a first user message that withholds the instruction and the remaining steps. \sysname therefore has no image of the scene yet. Its first implemented code, which follows an example in the system prompt, reads the instruction (``Pick up the banana and place it in the bowl'') and the remaining steps, lists the objects detected by perception, and shows itself two camera views, one annotated with the object identifiers and bounding boxes.

In turn~2, \sysname builds its own geometric model of the banana. It retrieves the banana's points, fits an oriented bounding box to them, and reads the end-effector pose and the table height from the robot's observation. In turn~3, \sysname plans the grasp. Because the banana is curved, \sysname reasons that the centroid of its points may not lie on it, so it plans to grasp the banana at the point of the cloud nearest to the centroid, with the fingers closing across the banana's width, perpendicular to its long axis. It plans timed waypoints for the approach and previews the motion without executing it. The harness confirms that the motion is feasible, and \sysname then runs a collision check on poses sampled along the previewed path, which reports contacts of the gripper with both the banana and the table. In turn~4, \sysname accepts the contact with the banana as expected for a grasp, raises the grasp pose slightly so that the fingertips reach approximately the middle of the banana's height rather than the table, and only then executes the approach. In turn~5, \sysname checks the wrist camera and finds that the banana sits between the open fingers. It closes the gripper and lifts the banana. In turn~6, \sysname interprets its gripper reading, which rose from 0.28 after closing to 0.55 after the lift, as the banana slipping or rotating in the grasp. Rather than reusing the bowl position from the first turn, it perceives the scene again to locate the bowl's center. It then plans to carry the banana high enough to clear the bowl's rim, and to lower it before opening the gripper. In turn~7, allowing for the banana having rotated in the grasp, it estimates that the banana now hangs about 7~cm below the gripper, so its bottom clears the rim during the carry. It then carries it over the bowl and lowers it. In turn~8, it opens the gripper and plans to raise the empty gripper to look at the result. Finally, RoboLab's success check detects that the banana has dropped into the bowl and ends the episode.

\begin{table*}[tp]
  \centering
  \footnotesize
  \setlength{\tabcolsep}{3pt}
  \caption{\textbf{Skills \sysname implements for itself.} Skills extracted from the \sysname run on RoboLab-120 (Opus~5.5, high reasoning effort, 1,200 episodes), grouped into scene reasoning, motion, and diagnostics. A skill is listed only if it appears in at least five episodes across at least two tasks.}
  \label{tab:skills}
  \begin{tabular}{|p{0.22\linewidth}|p{0.54\linewidth}|c|r|}
    \hline
    \textbf{Skill} & \textbf{Description} & \textbf{Code} & \shortstack[r]{\textbf{Episodes}\\\textbf{(of 1,200)}} \tabularnewline
    \hline
    \multicolumn{4}{|l|}{\textbf{Scene reasoning skills}} \tabularnewline
    \hline
    Measure container
      & \raggedright Measures a container's rim height, inner width, and floor height before placing an object in it.
      & \checkmark & 66\% (792) \tabularnewline
    Reason about object's axes
      & \raggedright Finds an object's long and short axes from its points, typically to grasp it across its short axis.
      & \checkmark & 47\% (560) \tabularnewline
    Find free placement space
      & \raggedright Finds a free space for each object so that placed objects do not overlap.
      & \checkmark & 40\% (476) \tabularnewline
    Ordering and stacking
    & \raggedright Plans the order of moves when stacking / handling several objects.
    & -- & 12\% (149) \tabularnewline
    Identify objects based on their category
      & \raggedright Decides which objects fit a category in the instruction.
      & -- & 7\% (82) \tabularnewline
    \hline
    \multicolumn{4}{|l|}{\textbf{Motion skills}} \tabularnewline
    \hline
    Align grasp with object's axes
    & \raggedright Sets the gripper's yaw based on the object's oriented bounding box or principal axes.
    & \checkmark & 61\% (731) \tabularnewline
    Grasp by handle / stem
    & \raggedright Grasps an object by its handle / stem.
    & -- & 34\% (410) \tabularnewline
    Side or tilted grasp
    & \raggedright Grasps from the side or at a tilt when a top down grasp is not possible, e.g.,\ for an object on a shelf.
    & \checkmark & 24\% (291) \tabularnewline
    Reorient or pivot
      & \raggedright Reorients an object into the desired pose, e.g.,\ pivots a cup upright.
      & \checkmark & 24\% (285) \tabularnewline
    Gently place
      & \raggedright Lowers the object until it nearly touches the target, then releases it.
      & \checkmark & 20\% (245) \tabularnewline
    Rim or wall grasp
    & \raggedright Grasps the rim or wall of an open container between one finger inside and one outside, e.g.,\ a bowl too wide to fit between the fingers.
    & \checkmark & 19\% (230) \tabularnewline
    Pause to settle
      & \raggedright Pauses after a motion or a release so the arm or object settles.
      & \checkmark & 17\% (207) \tabularnewline
    Drop into a container
      & \raggedright Releases the object above a deep container's opening, in an attempt to be clear of its walls.
      & \checkmark & 9\% (104) \tabularnewline
    Re-grasp
      & \raggedright Sets an object down and grasps it again for a better grasp, e.g.,\ after tipping a bottle.
      & \checkmark & 6\% (75) \tabularnewline
    Lift test
    & \raggedright Lifts an object a few centimeters and checks the gripper reading and cameras before carrying it.
    & \checkmark & 6\% (72) \tabularnewline
    Push
      & \raggedright Pushes an object it cannot grasp (too wide, or blocked) in a straight line.
      & \checkmark & 4\% (46) \tabularnewline
    Drag
      & \raggedright Slides an object, e.g.,\ a hammer, along a surface instead of lifting it.
      & \checkmark & 1\% (10) \tabularnewline
    \hline
    \multicolumn{4}{|l|}{\textbf{Diagnostic skills}} \tabularnewline
    \hline
    Repair a refused plan
      & \raggedright When the harness refuses waypoints (unreachable or too fast), it fixes the cause, e.g.,\ moves the target, rotates the gripper, or slows the motion, and previews again.
      & \checkmark & 40\% (481) \tabularnewline
    Diagnose a failed motion
      & \raggedright After a failed grasp or motion, checks the end-effector pose, gripper opening, motion status and cameras before re-planning.
      & \checkmark & 39\% (464) \tabularnewline
    Abandon a failed strategy
      & \raggedright Recognizes that a strategy has already failed and switches to another instead of retrying it.
      & -- & 12\% (143) \tabularnewline
    \hline
  \end{tabular}

  \vspace{3pt}
  \parbox{\linewidth}{\scriptsize
  Episodes: number of episodes (out of 1,200) in which the skill appears; detected from \sysname's code and/or thinking trace.
  Code: \checkmark{} in most episodes where the skill appears, \sysname implements and executes it; --~the skill appears only in its thinking trace.}
\end{table*}

\subsection{Case studies}
\label{sec:appendix:case_studies}

This section walks through a number of episodes in detail, to show how \sysname reasons, acts, and verifies its progress over the course of a task.

\subsubsection{Self-calibration of the wrist camera}
\label{sec:appendix:case_studies:self_calibration}

This case study follows one episode of the RGB-only ablation
(\cref{sec:results:analysis:perception}, \cref{fig:amp_ablation_results}C) on the RoboLab task \texttt{FoodPacking2CansTask} (``Pack canned foods into the bin''). In this setting, \sysname has access to RGB frames from the two scene cameras and the wrist camera, the robot's proprioception, and its kinematic model. The harness withholds depth, the camera intrinsics and extrinsics, and perception tools. \sysname can see the cans in every image, but it has nothing that turns a pixel into a position in the robot's frame, which it needs before it can plan a grasp.

\sysname closes this gap by calibrating the wrist camera itself. The wrist camera moves with the gripper, whose pose \sysname knows at every instant, so images taken at known gripper poses form a multi-view camera rig once two unknowns -- the camera's intrinsics and its fixed pose relative to the gripper -- are solved for. The episode succeeded in 44 model calls, of which the calibration and 3D estimation described below took the first 16, all before the first grasp.

\paragraph{Calibration.}
\sysname moved the gripper to a few known poses and kept the wrist image at each (\cref{fig:self_calib:views}A), then matched SIFT features across the views. It wrote a bundle adjustment routine that performs a least-squares fit of the camera's intrinsics, its pose relative to the gripper, and the 3D positions of the matched features. The bundle adjustment minimizes the distance between where each point projects and where it was observed. The first fit converged but placed the camera 2.8\,m up the tool axis, because shifts and a yaw alone leave the camera's offset along that axis unobservable. \sysname diagnosed this, added two views with the tool tilted
(\cref{fig:self_calib:views}B), and re-ran the fit on all seven views. The table plane emerged from the reconstruction without being given, and \sysname checked the result by projecting the
points back onto a view.

\begin{widefigure}
    \centering
    \begin{minipage}[t]{0.495\linewidth}
        \setlength{\unitlength}{\linewidth}%
        \begin{picture}(1,0.68246)
            \put(0,0){\includegraphics[width=\linewidth]{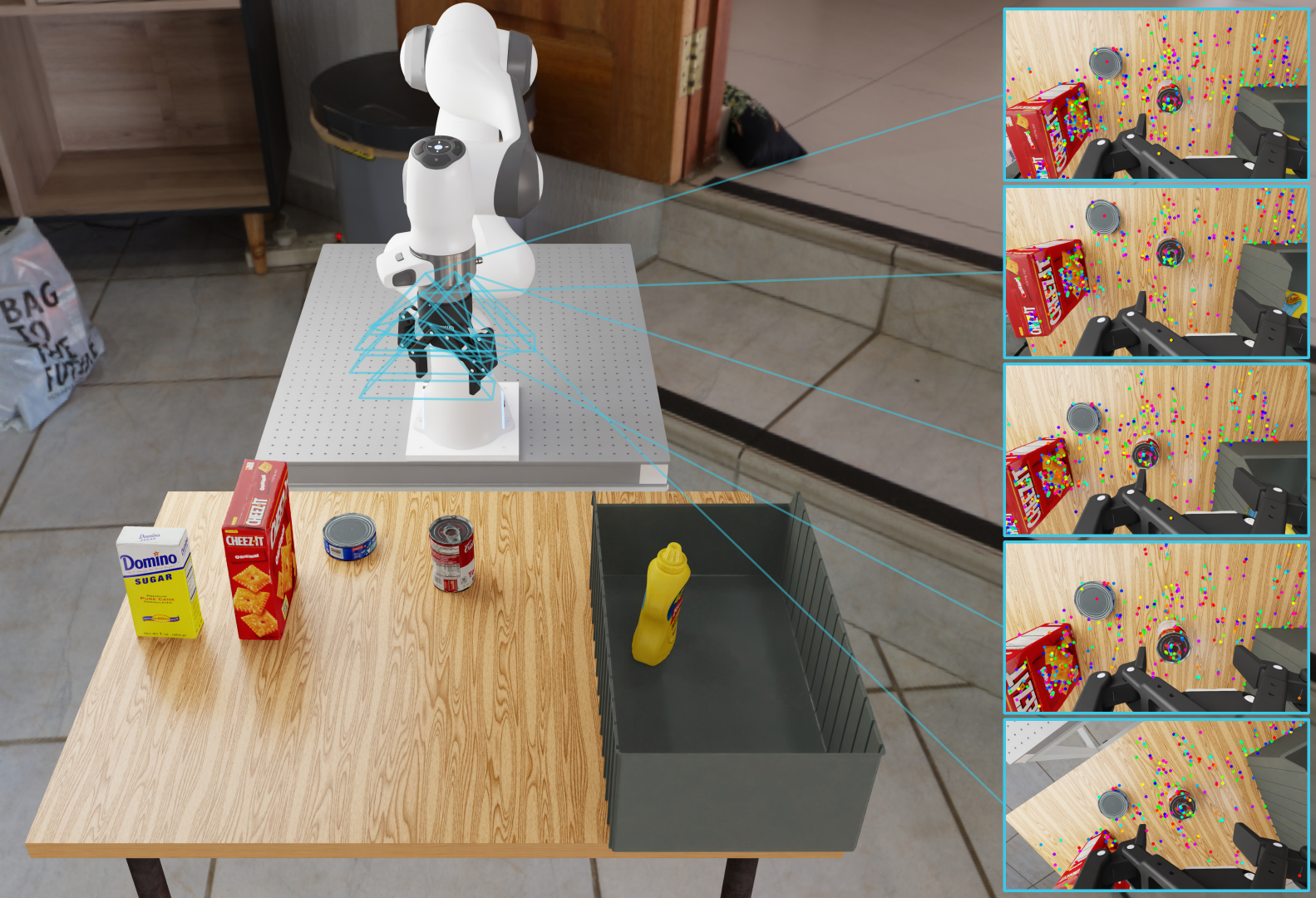}}
            \put(0.008,\lxtop{0.674}){\makebox(0,0)[tl]{\setlength{\fboxsep}{1pt}\colorbox{white}{\sffamily\bfseries\fontsize{9}{9}\selectfont A}}}
        \end{picture}
    \end{minipage}\hfill
    \begin{minipage}[t]{0.495\linewidth}
        \setlength{\unitlength}{\linewidth}%
        \begin{picture}(1,0.68246)
            \put(0,0){\includegraphics[width=\linewidth]{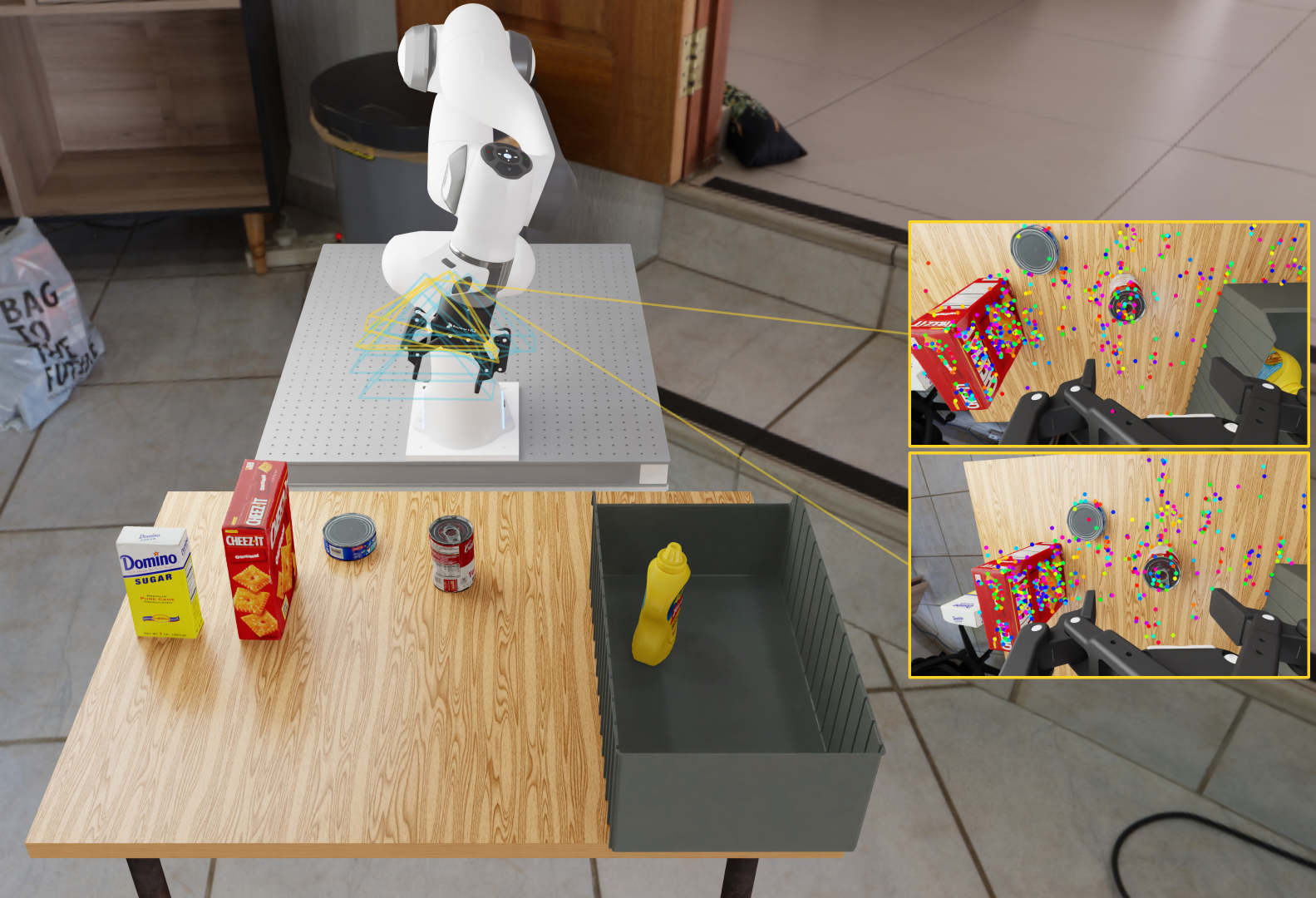}}
            \put(0.008,\lxtop{0.674}){\makebox(0,0)[tl]{\setlength{\fboxsep}{1pt}\colorbox{white}{\sffamily\bfseries\fontsize{9}{9}\selectfont B}}}
        \end{picture}
    \end{minipage}
    \caption{Wrist-camera views collected by \sysname to calibrate the camera, rendered from a fixed scene camera. Each frustum is the wrist camera's true pose where a view was taken. Dots mark SIFT features matched across the images. (\textbf{A}) First calibration round. Five views collected after known shifts, a descent and a yaw of the gripper. (\textbf{B}) Second calibration round. Two tilted views, added after \sysname found that the views in (\textbf{A}) leave the camera's offset along the tool axis unobservable.}
    \label{fig:self_calib:views}
\end{widefigure}

\paragraph{Triangulation.}
With the camera model, \sysname located the cans' top rims (\cref{fig:self_calib:triangulation}A) and the bin's rim and the mustard bottle inside it (\cref{fig:self_calib:triangulation}B) by intersecting rays from several views. The grasps and placements that followed used these positions directly.

\begin{widefigure}
    \centering
    \begin{minipage}[t]{0.495\linewidth}
        \setlength{\unitlength}{\linewidth}%
        \begin{picture}(1,0.68246)
            \put(0,0){\includegraphics[width=\linewidth]{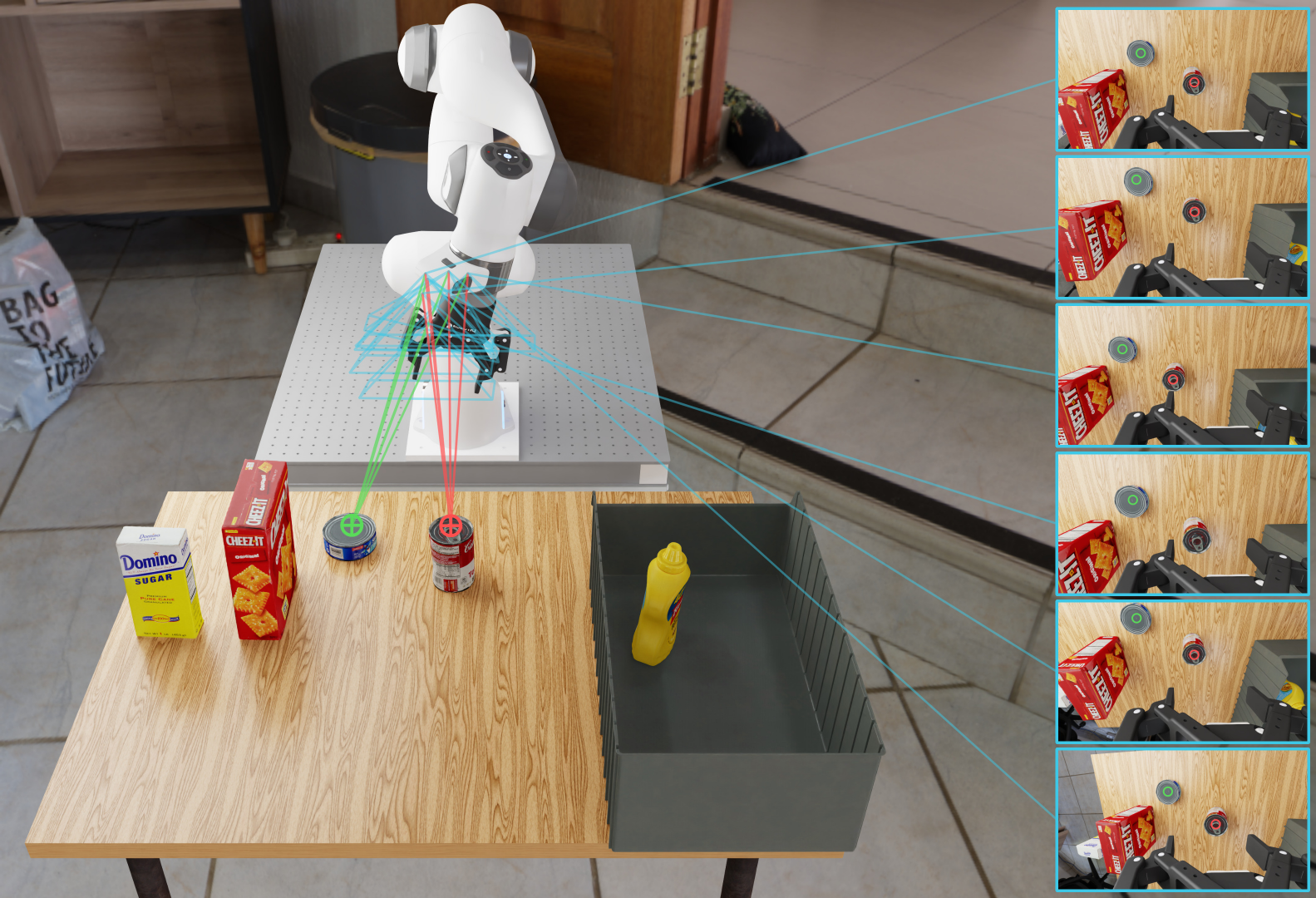}}
            \put(0.008,\lxtop{0.674}){\makebox(0,0)[tl]{\setlength{\fboxsep}{1pt}\colorbox{white}{\sffamily\bfseries\fontsize{9}{9}\selectfont A}}}
        \end{picture}
    \end{minipage}\hfill
    \begin{minipage}[t]{0.495\linewidth}
        \setlength{\unitlength}{\linewidth}%
        \begin{picture}(1,0.68246)
            \put(0,0){\includegraphics[width=\linewidth]{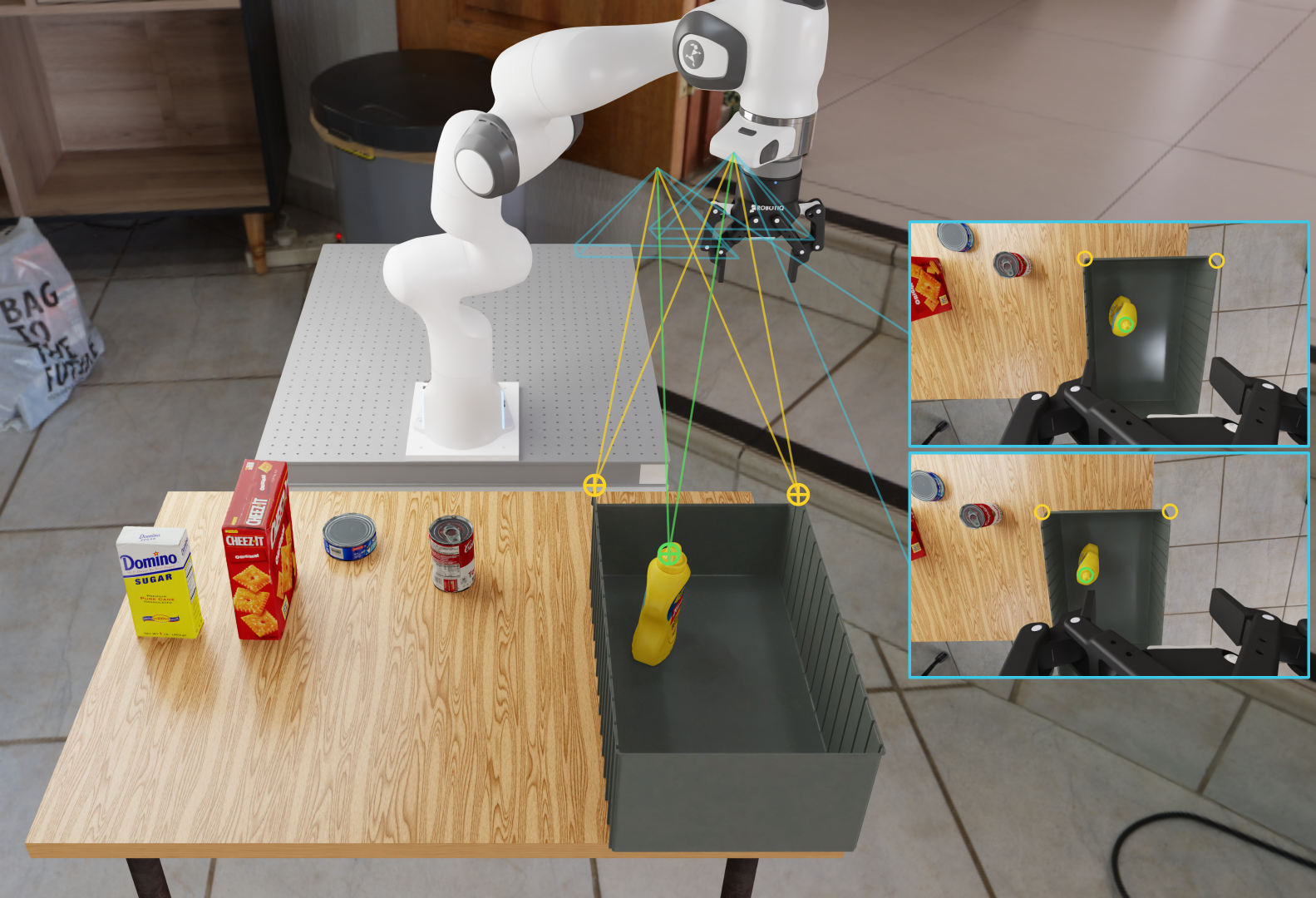}}
            \put(0.008,\lxtop{0.674}){\makebox(0,0)[tl]{\setlength{\fboxsep}{1pt}\colorbox{white}{\sffamily\bfseries\fontsize{9}{9}\selectfont B}}}
        \end{picture}
    \end{minipage}
    \caption{3D estimation from the self-calibrated wrist camera. Each frustum is a wrist view that \sysname triangulated from, linked to its image, in which \sysname's picked pixels are circled. Rays from the views meet at the triangulated points. (\textbf{A}) The top centers of the tuna can (green) and the soup can (red), from six calibration views. (\textbf{B}) The bin's two far rim corners (yellow) and the mustard bottle's cap (green), from two views over the bin.}
    \label{fig:self_calib:triangulation}
\end{widefigure}

\subsubsection{Active perception}
\label{sec:appendix:case_studies:active_perception}

This case study follows one episode of the ``Wrist camera only'' ablation
(\cref{sec:results:analysis:perception}, \cref{fig:amp_ablation_results}D) on the RoboLab task
\texttt{CubesAndBlocksInBinTask} (``Put all the cubes and blocks in the bin''). In this setting, the
wrist camera is \sysname's only view of the scene: its code receives the wrist camera's RGB and depth
images, and there is no scene camera. The task has seven target objects that the arm must pick up and
place in the bin: three Rubik's cubes and a blue, a red, a green, and a yellow block. Because the camera
rides on the gripper, it mostly sees only part of the table at a time. At the start, some targets are in view
and others are not, so \sysname first has to work out where to look.

\paragraph{Getting an overview.}
At the start pose, \sysname noted that there might be more objects beyond the edge of the frame and
hidden under the gripper (\cref{fig:active_perception}A), so it did not plan from this view.
Instead, it lifted the gripper to 0.62\,m above the middle of the table to see the whole scene. From
there it found the green block and the whole birdhouse. Two problems remained. The bin touched the
edge of the image, so its estimated box might be cut short, and objects might lie out of view on the
robot's side of the table. \sysname therefore moved the camera 0.20\,m toward the robot
(\cref{fig:active_perception}B). The bin's box grew from 34 to 42\,cm, and the yellow block
appeared next to a lizard figurine.

\paragraph{Taking a closer look.}
The overviews found every target but measured them poorly. The depth captured mostly
the tops of the objects from that height, so two of the Rubik's cubes and the blue block came out about 2\,cm tall instead of about 5\,cm. The agent reasoned that perception from that height was too coarse and decided to hover low over each object and look again before grasping it. At the first close look, the Rubik's cube nearest the robot
measured 6.1$\times$6.1$\times$5.3\,cm, with its faces at $-51.6$\textdegree{} rather than the
69.9\textdegree{} the overview had given. \sysname then used the new angle to choose how to close its
fingers so that the gripper would not touch the blue block beside the cube (\cref{fig:active_perception}C).
For every later pick, \sysname started with a close look like this one.

\begin{widefigure} %
	\centering
    	\includegraphics[width=\linewidth, trim=0pt 0pt 0pt 0pt, clip]{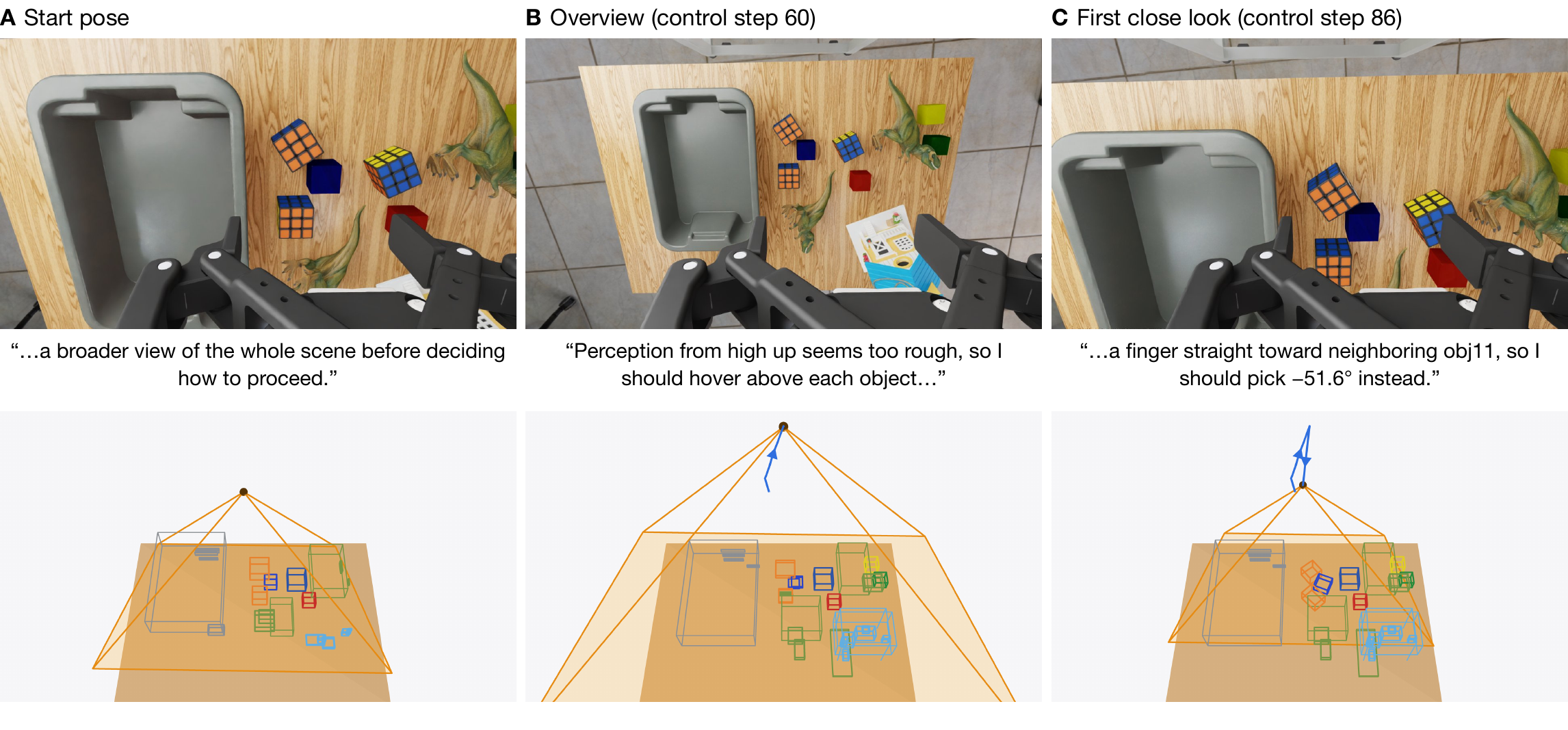}

	\caption{\textbf{Scene perception with only a wrist camera.} The task is “Put all the cubes and blocks in the bin.” Each column is one stage before \sysname decides to grasp the first target object: (\textbf{A}) the start pose, (\textbf{B}) the overview at control step 60, and (\textbf{C}) the first close look at control step 86. The top row is what the wrist camera sees, with a quote from \sysname's thinking trace at that view. The bottom row shows the same moment in 3D space: the camera (dark dot), the space it can see, the path it has taken so far (blue, with arrows for the direction), and the objects \sysname's perception tools have reported by that point. In (\textbf{A}), only part of the table is in view, so \sysname raises the camera for a broader view. In (\textbf{B}), all seven targets and the whole bin are perceived, but their boxes are rough, so the agent decides to hover above each object before grasping it. In (\textbf{C}), objects are measured again after the gripper hovers 0.30 m above the table. \sysname uses the cube's newly measured orientation to choose a closing angle that keeps a finger off the neighboring blue block.}
	\label{fig:active_perception}
\end{widefigure}

\subsection{Robustness to instruction phrasing on RoboLab-120}
\label{sec:appendix:instruction_spread}
\begin{figure}[t]
    \centering
    \includegraphics[width=\linewidth]{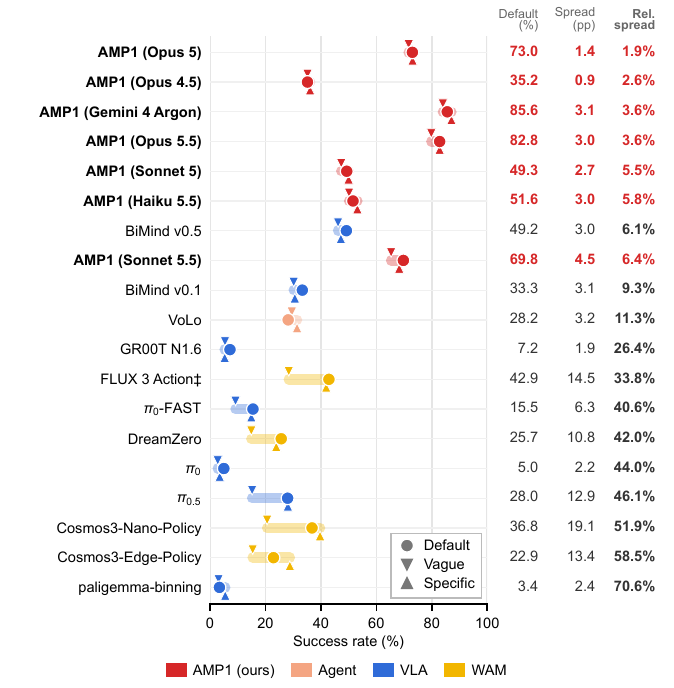}
    \caption{\textbf{Robustness to instruction phrasing on RoboLab-120.}
    Success rate across the \emph{default} (circle), \emph{vague} ($\blacktriangledown$), and \emph{specific} ($\blacktriangle$) instruction conditions for all seven \sysname configurations and 12 baseline policies evaluated across all three conditions, ranked by relative spread from lowest (most robust) to highest. Shaded bars span the range from minimum to maximum success rate across the three conditions. Columns on the right report each policy's default success rate (\%), absolute spread ($\max - \min$, in percentage points), and relative spread ($\text{spread} / \text{default success rate}$). $^{\ddagger}$Evaluated by us using the released checkpoint on the vague and specific instruction conditions (1,200 trials per condition), with default from the leaderboard.}
    \label{fig:instruction_spread}
\end{figure}

RoboLab-120 evaluates every task under three natural-language instruction conditions: \emph{default} (standard task phrasing), \emph{vague} (underspecified phrasing that omits explicit object names or attributes), and \emph{specific} (detailed phrasing that names target objects and relations explicitly)~\cite{robolab}. \cref{fig:instruction_spread} compares all 19 policies for which evaluation data exist across all three conditions: seven \sysname configurations (Gemini~4 Argon, Opus~5.5, Opus~5, Sonnet~5.5, Haiku~5.5, Sonnet~5, and Opus~4.5) and 12 baseline policies (11 leaderboard entries plus our evaluation of FLUX~3 Action~\cite{flux3a} on the vague and specific conditions, 1,200 trials per condition). For each policy, we report the absolute success rate under default instructions, the absolute spread across the three conditions ($\max - \min$, in percentage points), and the relative spread ($\text{spread} / \text{default success rate}$), and rank all policies from lowest to highest relative spread.
Across all seven models, \sysname's absolute spread stays between 0.9 and 4.5 percentage points, corresponding to a relative spread of 1.9\% to 6.4\%~-- occupying the top six positions and seven of the top eight. \sysname with six models (Opus~5, Opus~4.5, Gemini~4 Argon, Opus~5.5, Sonnet~5, and Haiku~5.5) achieves the six lowest relative spreads on the benchmark (1.9\%, 2.6\%, 3.6\%, 3.6\%, 5.5\%, and 5.8\%, respectively). Even with smaller or older models such as Haiku~5.5 (51.6\% default success) and Opus~4.5 (35.2\% default success), \sysname maintains relative spreads of 5.8\% and 2.6\%, lower than every learned policy. Only one learned policy, BiMind~v0.5 (6.1\%), has a lower relative spread than any \sysname configuration: \sysname with Sonnet~5.5 (6.4\%), a gap of 0.3 points. By contrast, end-to-end VLA and WAM/LWM baselines exhibit a median relative spread of 42.0\%, ranging from 6.1\% for BiMind~v0.5 to 70.6\% for paligemma-binning, with large drops under vague phrasing~-- including 12.8 percentage points for $\pi_{0.5}$~\cite{pi05} (46.1\% relative spread), 14.5 percentage points for FLUX~3 Action~\cite{flux3a} (33.8\%), and 16.2 percentage points for Cosmos3-Nano-Policy (51.9\%).

\subsection{RoboLab-120 failure mode analysis}
\label{sec:appendix:robolab_failures}
We examined every failed episode of the two strongest models in the RoboLab-120 evaluation of \cref{fig:simeval}B, under the `default' instruction condition (10 trials per task, 1,200 trials per model): 173 failed episodes for \sysname with Gemini 4 Argon (1,027 successes, or 85.6\%) and 206 for \sysname with Opus 5.5 (994 successes, or 82.8\%). For each failed episode, we compared the final scene with RoboLab's programmed verifier and analyzed the agent's execution trace and measurements, the task verification code, and the final camera frames.

\cref{fig:failure_sankey} partitions all 1,200 trials of each model by outcome. It first separates failures by their source~-- the benchmark (\emph{Eval failure}), the policy (\emph{\sysname failure}), or the model API (\emph{Provider failure})~-- and then breaks each source into specific causes. \sysname succeeds in every trial of 76 of the 120 tasks with Gemini 4 Argon (66 tasks with Opus 5.5). For both models, roughly two-thirds of the failures stem from the benchmark rather than the policy: 121 of the 173 failures with Gemini 4 Argon (70\%), across 25 tasks, and 135 of the 206 failures with Opus 5.5 (66\%), across 30 tasks. These benchmark artifacts comprise objects whose width or mass distribution is incompatible with a standard parallel-jaw grasp (77 episodes for Gemini 4 Argon and 101 for Opus 5.5) and flaky verifiers that reject end states that satisfy the instruction (44 and 34 episodes, respectively). Genuine \sysname failures on graspable objects account for 47 episodes with Gemini 4 Argon (3.9\% of all trials) and 71 with Opus 5.5 (5.9\%); the remaining 5 episodes with Gemini 4 Argon are terminated by the provider's safety filter.

\begin{widefigure}
    \centering
    \includegraphics[width=\linewidth]{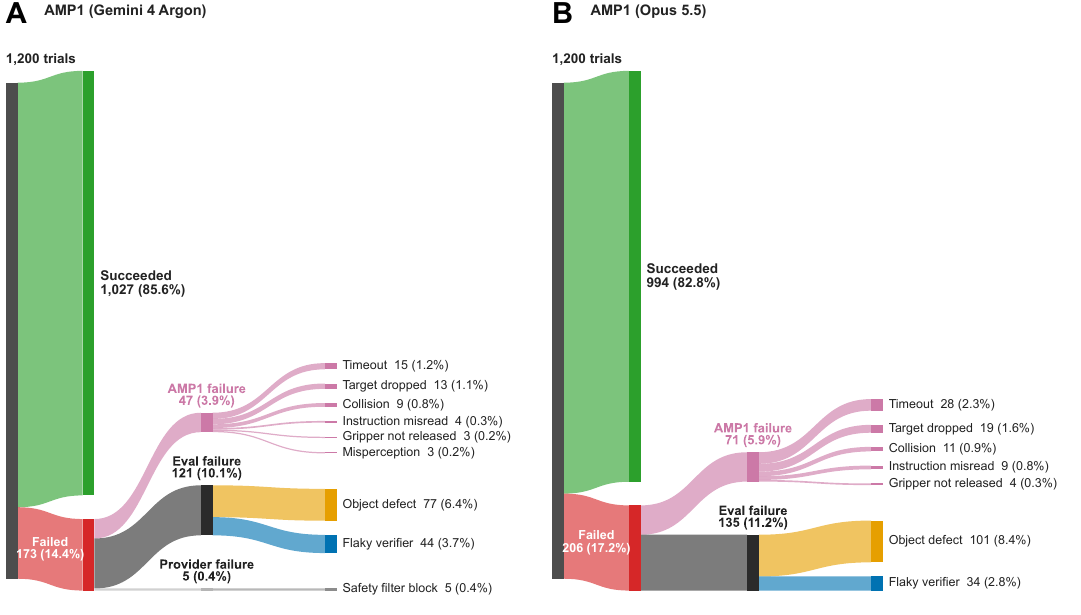}
    \caption{\textbf{Outcome and failure breakdown across all 1,200 RoboLab-120 trials.}
    (\textbf{A}) \sysname with Gemini 4 Argon (1,027 succeeded, 173 failed).
    (\textbf{B}) \sysname with Opus 5.5 (994 succeeded, 206 failed).
    In each panel, all 1,200 trials in the `default' instruction condition (10 trials across 120 tasks) branch into succeeded and failed trials; failed trials then branch by whether the failure originates in the policy (\emph{\sysname failure}), the benchmark (\emph{Eval failure}), or the model API (\emph{Provider failure}), and finally into specific causes. Band widths are proportional to the number of trials at a single scale shared across (\textbf{A}) and (\textbf{B}). Roughly two-thirds of all failed trials~-- 121 of 173 for Gemini 4 Argon (70\%) and 135 of 206 for Opus 5.5 (66\%)~-- are evaluation failures caused by objects that are incompatible with a standard parallel-jaw grasp (\emph{Object defect}) or by flawed verifiers that reject valid end states (\emph{Flaky verifier}).}
    \label{fig:failure_sankey}
    \label{tab:failure_tiers}
\end{widefigure}

\mypara{Eval failure: Object defect (77 episodes for Gemini 4 Argon, 101 for Opus 5.5)}
In 16 tasks for Gemini 4 Argon and 22 tasks for Opus 5.5 (24 distinct tasks in total; \cref{fig:failure_sankey}), the target object is incompatible with a standard parallel-jaw grasp. Some objects, such as the pomegranate (115~mm), the bagels (100~mm), and the coffee can (102~mm), are wider than the 85-mm maximum aperture of the Robotiq 2F-85 gripper. Others slip out of a friction grasp because of their mass distribution or thin profile, including the 1.5-kg cordless drill, heavy-headed hammers extracted from a bin, a 2-cm-thick flat block, and a moka pot. Only one task fails every trial for both models because of a defect: in \texttt{FoodPackingByColor}, the instruction asks for ``the blue object'' although the scene contains two, and RoboLab's verifier accepts only the 102-mm coffee can. In all 10 trials, \sysname packs the graspable blue SPAM can and fails to hold the oversize coffee can. In the remaining tasks, \sysname succeeds when it improvises a non-standard grasp (\cref{sec:results:emergent_behaviors}). For objects wider than the gripper, it holds a narrower feature or supports the object from below: it inserts one finger through the hole of a bagel (\texttt{BagelsOnPlate} and \texttt{GrabABagel}), and it pinches the pomegranate's narrow crown or pins the fruit against a heavier object and scoops it up with the open gripper (\texttt{FruitsOnPlate} and \texttt{ClearOrganicObjects}; \cref{fig:emergent_behavior}D). For the heavy drill, it replaces the friction grasp with an interlocking one, closing the fingers around the top of the handle so that the head rests on them once the drill is lifted (\texttt{PickDrill} and \texttt{ToolsPickingDrill}; \cref{fig:emergent_behavior_appendix_2}C). Across the 16 affected tasks (160 trials), \sysname with Gemini 4 Argon still succeeds in 66 episodes (41\%); it fails in 77 because of the object defect and in 17 for other reasons (5 flaky-verifier rejections, 8 \sysname errors, and 4 provider blocks). Across the 22 tasks affected for \sysname with Opus 5.5 (220 trials), it succeeds in 107 episodes (49\%).

\mypara{Eval failure: Flaky verifier (44 episodes for Gemini 4 Argon, 34 for Opus 5.5)}
In 13 tasks for Gemini 4 Argon and 9 tasks for Opus 5.5 (\cref{fig:failure_sankey}), RoboLab's verifier rejects end states that satisfy the natural-language instruction. Only one task fails every trial for both models: in \texttt{RecycleCartonsOnBox} (``Put the cartons that can be recycled on the box''), the verifier tests whether the cartons lie \emph{inside} the box, which is closed, whereas the task's own progress tracker checks whether they rest on top of it. In all 10 trials for both models, \sysname places both cartons on top of the box, the progress tracker scores them as placed, and the verifier rejects the outcome.
In the remaining tasks, the verifier is overly strict rather than impossible to satisfy: it accepts only a narrow subset of valid placements, so \sysname still succeeds in 59 of the 130 trials across the 13 tasks with Gemini 4 Argon and in 48 of the 90 trials across the 9 tasks with Opus 5.5. In \texttt{CookingPickPastaTool} (9 verifier rejections for Gemini 4 Argon and 7 for Opus 5.5), the pasta tool ends standing inside the target holder, but the verifier~-- which requires the tool's center to lie inside the holder while the tool touches it~-- does not register it; the single Gemini 4 Argon success came when the episode ended with the gripper still holding the tool inside the holder. In \texttt{PickUpBluePitcher} (6 and 5 rejections), \sysname lifts the pitcher 8.5 to 12.8~cm above the table, but the verifier's 5-cm lift check does not trigger. In \texttt{StackWhiteMugs} (5 rejections for Gemini 4 Argon), the verifier rejects deeply nested mugs whose upper bounding-box corners dip below the lower mug's base; in \texttt{PickOrangeObject} (5 for Opus 5.5), it rejects lifts that sit on the boundary of its height threshold. In \texttt{MoveBananaToBagelPlate} (5 rejections for Gemini 4 Argon), \sysname places the banana on top of the bagels already on the plate, whereas the verifier requires direct contact with the plate surface. The remaining rejections (9 across 8 tasks for Gemini 4 Argon and 7 across 5 tasks for Opus 5.5) are isolated cases of one or two episodes in which a lifted object or an in-container placement sits just outside the verifier's geometric threshold, as in \texttt{PickDrill}, \texttt{GrabABagel}, \texttt{UtensilsInMug}, and \texttt{ToolsPickingAllHammers}. In nearly all of these episodes, \sysname ended the trial believing it had succeeded.

\mypara{\sysname failures and provider blocks (52 episodes for Gemini 4 Argon, 71 for Opus 5.5)}
The remaining failures are execution or interpretation errors by \sysname on objects that the gripper can hold with a standard grasp: 47 episodes across 24 tasks for Gemini 4 Argon and 71 episodes across 33 tasks for Opus 5.5 (\cref{fig:failure_sankey}).
\begin{itemize}
    \item \emph{Timeout (15 episodes for Gemini 4 Argon, 28 for Opus 5.5):} \sysname exhausts the step budget, or ends the episode itself because too few steps remain to finish the remaining objects. Most timeouts follow earlier setbacks that \sysname recovers from~-- repeated regrasps after slips, freeing a wedged finger, or searching for a reachable arm configuration on a shelf~-- but that consume most of the step budget before the last object is placed.
    \item \emph{Target dropped (13 for Gemini 4 Argon, 19 for Opus 5.5):} A graspable object slips, drops, or tips out of a valid goal state during transport or release, and \sysname does not recover it before the episode ends.
    \item \emph{Collision (9 and 11 episodes):} The arm, the gripper, or the held object collides with the scene, such as an unperceived upper shelf tier or a bin wall, or knocks a previously placed object out of position.
    \item \emph{Instruction misread (4 for Gemini 4 Argon, 9 for Opus 5.5):} \sysname identifies the objects in the scene correctly but interprets the instruction too narrowly or too broadly. In \texttt{SpoonsInPot} (4 episodes for Gemini 4 Argon and 7 for Opus 5.5) and \texttt{GreenSpoonsInPot} (1 for Opus 5.5), it takes a deep green utensil with a bent neck for a soup ladle rather than a serving spoon and leaves it out (in one trial actively removing it from the pot), whereas RoboLab's verifier requires it in the pot. In one trial of \texttt{OneBottleInSquarePail}, Opus 5.5 puts a white-capped utility jug into the pail instead of one of the small white plastic bottles.
    \item \emph{Gripper not released (3 for Gemini 4 Argon, 4 for Opus 5.5):} The object reaches the goal region, but the gripper fails to open or detach cleanly, for example when a bowl rim or a mug handle wedges on the finger linkage, and remains in contact at the end of the episode.
    \item \emph{Misperception (3 for Gemini 4 Argon, 0 for Opus 5.5):} \sysname misidentifies an object's category or geometry from the camera images. In two trials of \texttt{SpoonsInPot}, Gemini 4 Argon takes an oval green serving spoon, partly hidden beneath the ladle, for a flat spatula and leaves it on the plate; in one trial of \texttt{FoodPacking1Boxes}, it mistakes a crease on top of a cracker box for the box's edge and misplaces the grasp.
\end{itemize}
Finally, 5 episodes with Gemini 4 Argon (\emph{Provider failure: Safety filter block}) end early because the Gemini API blocks a model request under its safety filter, even after automatic retries; no such blocks occur with Opus 5.5.

\mypara{Summary} On the tasks free of both flaky verifiers and object defects, \sysname succeeds in 96.2\% of episodes with Gemini 4 Argon (914 of 950 trials across 95 tasks) and in 93.6\% with Opus 5.5 (842 of 900 trials across 90 tasks). On the 84 tasks free of both issues for both models, the success rates are 96.2\% and 93.7\%, respectively. Without excluding any task, crediting the verifier rejections whose end state satisfies the instruction (44 for Gemini 4 Argon and 34 for Opus 5.5) raises \sysname's success rate across all 1,200 trials from 85.6\% to 89.3\% with Gemini 4 Argon and from 82.8\% to 85.7\% with Opus 5.5.

\subsection{Inference cost of \sysname}
\label{sec:appendix:ablation_cost}

This section estimates the inference cost of \sysname for the models with public API pricing, the Claude models and Gemini~4 Argon, by pricing each episode's tokens at list price (\cref{tab:pricing}). With Opus~5.5, a RoboLab-120 episode costs \$1.39 on average, about 26 times lower than the \$36.40 reported for Astranomy with GPT-6 Astra~\cite{yang2026astranomy}. (\sysname also succeeds more often, 82.8\% against 74.8\%.) Astranomy queries the model for every small end-effector step, whereas a single \sysname turn can command a whole trajectory through code, and images enter the model's context only when it asks for them.

Cost per episode is widely reported, but it mixes two factors: the tokens an episode consumes, which the model and harness determine, and the price per token, which the provider sets. Prices do not track capability. Opus~5 is billed at 25\% more than Opus~5.5 for input and output tokens and 2.5 times more for cache reads (\cref{tab:pricing}). Prices also fall quickly: the cost of a given capability level has dropped by roughly an order of magnitude per year~\cite{Gundlach2025,EmbersonRoodman2026}. We therefore report model calls and token counts alongside cost (\cref{fig:ablation_cost}); unlike cost, they compare models and harnesses independently of the price list.

\mypara{Reasoning} More reasoning raises the success rate of every model in \cref{fig:ablation_cost}A, but it also substantially raises the cost per episode. The extra cost has two sources: the model writes more in each turn, and it takes more turns, each of which re-reads the episode (\cref{fig:ablation_cost}B, C). How much the extra cost buys depends on the model. For Opus~5.5 and Sonnet~5.5, more reasoning adds 14.6 and 17.2 points for about \$0.90 more per episode. For Gemini~4 Argon, which has little headroom, it adds 2.3 points for \$1.30 more, more than doubling the cost.

\mypara{Harness} For the weaker models (Haiku~5.5, Sonnet~5, and Opus~4.5), the assistive harness is both more successful and cheaper than the basic harness; for the stronger models (Gemini~4 Argon, Opus~5.5, Sonnet~5.5, and Opus~5), it is less successful at about the same cost (\cref{fig:ablation_cost}D; see \cref{sec:results:analysis:motion}). The weaker models take fewer turns, as the assistive routines replace much of their trial and error, and require fewer tokens (\cref{fig:ablation_cost}E, F). Sonnet~5 and Opus~4.5 take about 30\% fewer turns and cost 36 to 39\% less per episode; Haiku~5.5 costs about 10\% less. The stronger models take about as many turns as with the basic harness and generate 7 to 10\% fewer tokens, but they read 10 to 19\% more input, so their cost rises by up to 7\%.

\begin{table}[tp]
  \centering
  \caption{\textbf{List prices of a number of models.} (\$ per million tokens). Thinking is billed as output. Gemini~4 Argon has no separate cache-write price; its first pass is billed as input.}
  \label{tab:pricing}
   \ifcsname iflatexml\endcsname\small\fi\begin{tabular}{lcccc}
    \toprule
    Model & Input & Cache write & Cache read & Output \\
    \midrule
    Gemini~4 Argon & 4    & 4     & 0.2  & 20  \\
    Opus~5.5       & 4    & 5     & 0.2  & 20  \\
    Sonnet~5.5     & 2    & 2.5   & 0.1  & 10  \\
    Haiku~5.5      & 0.10 & 0.125 & 0.01 & 0.5 \\
    \midrule
    Opus~5         & 5    & 6.25  & 0.5  & 25  \\
    Opus~4.5       & 5    & 6.25  & 0.5  & 25  \\
    Sonnet~5       & 2    & 2.5   & 0.2  & 10  \\
    \bottomrule
  \end{tabular}
\end{table}

\begin{widefigure}
    \centering
    \includegraphics[width=\linewidth]{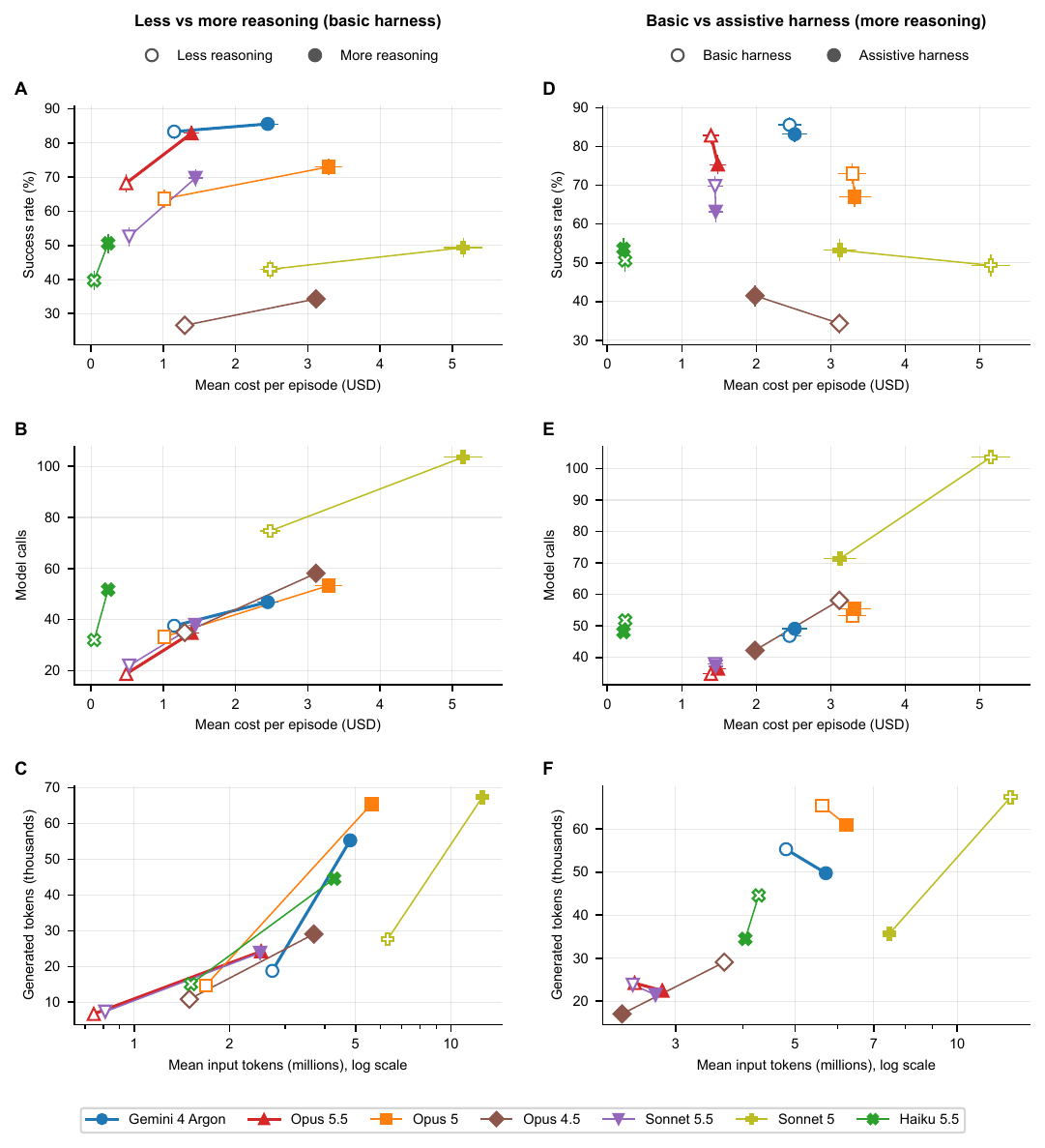}
    \caption{\textbf{Cost, model calls, and token use on RoboLab-120.} (Per-episode averages.)
    Left column: less vs.\ more reasoning (basic harness). Right column: basic vs.\ assistive harness (more reasoning).
    (\textbf{A}, \textbf{D}) Success rate vs.\ cost.
    (\textbf{B}, \textbf{E}) Model calls vs.\ cost.
    (\textbf{C}, \textbf{F}) Generated (output and thinking) vs.\ input tokens (uncached, cache write, and cache read; log scale).
    Error bars on cost and success rate are 95\% confidence intervals. Costs at list price (\cref{tab:pricing}).} 
    \label{fig:ablation_cost}
\end{widefigure}

\end{document}